\documentclass[10pt,journal,compsoc]{IEEEtran}

\usepackage[frozencache,cachedir=.]{minted}
\usepackage{amsfonts}
\usepackage{multicol}
\usepackage{multirow}
\usepackage{threeparttable}

\ifCLASSOPTIONcompsoc
  \usepackage[nocompress]{cite}
\else
  \usepackage{cite}
\fi
\ifCLASSINFOpdf
   \usepackage[pdftex]{graphicx}
   \graphicspath{{../pdf/}{../jpeg/}}
   \DeclareGraphicsExtensions{.pdf,.jpeg,.png}
\else
\fi
\usepackage{amsmath}

\usepackage[ruled,linesnumbered]{algorithm2e}
\makeatletter
\newcommand{\removelatexerror}{\let\@latex@error\@gobble}
\makeatother

\usepackage{array}

\usepackage{mdwmath}
\usepackage{mdwtab}

\ifCLASSOPTIONcompsoc
 \usepackage[caption=false,font=footnotesize,labelfont=sf,textfont=sf]{subfig}
\else
 \usepackage[caption=false,font=footnotesize]{subfig}
\fi

\ifCLASSOPTIONcaptionsoff
 \usepackage[nomarkers]{endfloat}
\let\MYoriglatexcaption\caption
\renewcommand{\caption}[2][\relax]{\MYoriglatexcaption[#2]{#2}}
\fi
\usepackage{url}
\newcommand\MYhyperrefoptions{bookmarks=true,bookmarksnumbered=true,
pdfpagemode={UseOutlines},plainpages=false,pdfpagelabels=true,
colorlinks=true,linkcolor={black},citecolor={black},urlcolor={black},
pdftitle={Merak: An Efficient Distributed DNN Training Framework with Automated 3D Parallelism for Giant Foundation Models},
pdfsubject={Merak},
pdfauthor={Zhiquan Lai, Shengwei Li, Xudong Tang, Keshi Ge, Weijie Liu, Yabo Duan, Linbo Qiao, and Dongsheng Li},
pdfkeywords={Distributed Systems, Deep learning, Foundation model training
}}
\ifCLASSINFOpdf
\usepackage[\MYhyperrefoptions,pdftex]{hyperref}
\else
\usepackage[\MYhyperrefoptions,breaklinks=true,dvips]{hyperref}
\usepackage{breakurl}
\fi
\begin{document}

%
\title{Merak: An Efficient Distributed DNN Training Framework with Automated 3D Parallelism for Giant Foundation Models}
%
%
%
%

\author{Zhiquan~Lai, 
Shengwei~Li, 
Xudong~Tang, 
Keshi~Ge, 
Weijie~Liu, 
Yabo~Duan, 
Linbo~Qiao, 
and~Dongsheng~Li 
\IEEEcompsocitemizethanks{\IEEEcompsocthanksitem Zhiquan~Lai, Shengwei~Li, Xudong~Tang, Keshi~Ge, Weijie~Liu, Yabo~Duan, Linbo~Qiao, and~Dongsheng~Li are with the National Laboratory for Parallel and Distributed Processing, College Of Computer, National University of Defense Technology in Changsha, Hunan, China \protect\\
E-mail: \{zqlai,gekeshi,liuweijie,yaboduan,qiao.linbo,dsli\}@nudt.edu.cn, \{lucasleesw9,txacs1993\}@gmail.com
\IEEEcompsocthanksitem The first two authors contributed equally to this work.}
}

\IEEEtitleabstractindextext{%
\begin{abstract}
Foundation models are in the process of becoming the dominant deep learning technology. Pretraining a foundation model is always time-consuming due to the large scale of both the model parameter and training dataset. Besides being computing-intensive, the pretraining process is extremely memory- and communication-intensive. These challenges make it necessary to apply 3D parallelism, which integrates data parallelism, pipeline model parallelism, and tensor model parallelism, to achieve high training efficiency.

However, current 3D parallelism frameworks still encounter two issues: i) they are not transparent to model developers, requiring manual model modification to parallelize training, and ii) their utilization of computation resources, GPU memory, and network bandwidth is insufficient. We propose \textit{Merak}, an automated 3D parallelism deep learning training framework with high resource utilization. Merak automatically deploys 3D parallelism with an automatic model partitioner, which includes a graph-sharding algorithm and proxy node-based model graph. Merak also offers a non-intrusive API to scale out foundation model training with minimal code modification. In addition, we design a high-performance 3D parallel runtime engine that employs several techniques to exploit available training resources, including a shifted critical path pipeline schedule that increases computation utilization, stage-aware recomputation that makes use of idle worker memory, and sub-pipelined tensor model parallelism that overlaps communication and computation. 

Experiments on 64 GPUs demonstrate Merak's capability to speed up training performance over state-of-the-art 3D parallelism frameworks of models with 1.5, 2.5, 8.3, and 20 billion parameters by up to 1.42, 1.39, 1.43, and 1.61$\times$, respectively. The code for Merak has been open-sourced at \url{https://github.com/hpdl-group/Merak}.
\end{abstract}

\begin{IEEEkeywords}
Distributed Systems, Deep learning, Foundation model training
\end{IEEEkeywords}}

\maketitle

\IEEEdisplaynontitleabstractindextext

%
\IEEEpeerreviewmaketitle

\ifCLASSOPTIONcompsoc
\IEEEraisesectionheading{\section{Introduction}\label{sec:introduction}}
\else
\section{Introduction}
\label{sec:introduction}
\fi

\IEEEPARstart{F}{}oundation deep neural networks (DNNs)~\cite{foundationmodel} have advanced rapidly in recent years.
State-of-the-art (SOTA) foundation DNNs possess high-dimensional features and have achieved promising results across various fields.

However, the training of foundation models presents challenges in terms of the scale of parameters and datasets. 
The giant model size and high dataset volume of foundation DNNs incur extremely high computation times, and it is foreseeable that the model size will continuously increase in the future.
These requirements necessitate \emph{3D parallelism}, which integrates data parallelism (DP), tensor model parallelism (TMP), and pipeline model parallelism (PMP) into a multi-node distributed training system. 
TMP and PMP are developed to address the issue associated with giant models. 
TMP splits the parameters in a model layer over multiple GPUs, whereas PMP partitions the model layers into stages and executes each stage in a pipelined manner. 
And DP deploys multiple model replicas to accelerate large dataset processing. 
Current 3D parallelism systems successfully train representative transformer-based~\cite{transformer} models at scale, such as 
GPT-3~\cite{brown2020language}, MT-NLG~\cite{mt-nlg}, and PanGu-$\alpha$~\cite{mindspore}.

\begin{table}[tb]
    \renewcommand{\arraystretch}{1.3}
    \centering
    \caption{The automatic parallel strategy availability on PyTorch based distributed training frameworks. $\times$ means needing model code rewriting.}
    \label{tab:autocompare}
    \begin{threeparttable}
    \begin{tabular}{c c c c}
    \hline
    System  & Auto DP & Auto PMP & Auto TMP \\
    \hline
    DeepSpeed~\cite{ds3d}& \checkmark & $\times$ & $\times$ \\
    \hline
    Megatron-LM~\cite{megatronsc} & \checkmark & $\times$ & $\times$ \\
    \hline
    Colossal-AI~\cite{colossal}  & \checkmark & $\times$ & $\times$ \\
    \hline
    Varuna~\cite{varuna} & \checkmark  & \checkmark* & - \\
    \hline
    SageMaker~\cite{karakus2021amazon}  & \checkmark  & \checkmark & $\times$ \\
    \hline
    \emph{Merak}   & \checkmark & \checkmark & \checkmark \\
    \hline
    \end{tabular}
\begin{tablenotes}
\footnotesize
\item[*]Varuna does not support TMP and requires adding CutPoints in model manually for PMP.
\end{tablenotes}
\end{threeparttable}
\end{table}

However, existing 3D parallelism systems still have two drawbacks, the first being \textbf{lack of generality}. 
DNN training frameworks~\cite{oneflow,mindspore}, such as PaddlePaddle~\cite{paddle}, integrate 3D parallelism as their training strategy for giant models, but these frameworks need experienced developers.
In PyTorch~\cite{torch}, one of the most commonly used DNN frameworks with high reputation for friendly usage, 
existing 3D parallelism libraries require system expertise and manual code refactoring, which further complicates the training process.
We summarize the availability of parallel strategies among PyTorch-based libraries in Table~\ref{tab:autocompare}. 
As the TMP customizes operators, little work can escape from manual module settings. 
For PMP, DeepSpeed~\cite{ds3d} and Megatron-LM~\cite{megatronsc} 
demand sequential models consisting of a single flat stack of layers.
Colossal-AI~\cite{colossal} requires redefining models with its particular interface. Varuna~\cite{varuna} asks the addition of specific CutPoint instances to model code. And SageMaker~\cite{karakus2021amazon} declares that it can apply PMP to any model, but it is proprietary and only available on AWS.

Researchers often share the latest progress in model training and architectures. Plentiful models and corresponding local training scripts exist in communities such as the open-source library Transformers~\cite{Wolf_Transformers_State-of-the-Art_Natural_2020}. Users can build, train, and deploy SOTA models without accounting for training and model details.
The complex code refactoring of existing 3D parallelism frameworks keeps us from the convenience of these community models, presenting a barrier in the development of the AI community.
Therefore, our first motivation is to enable general \emph{3D parallelism without modifying the original models}.

Another issue with 3D parallelism is \textbf{inefficiency}.
There are three main types of training resources for each worker: computation, memory, and network bandwidth. 
We observe 
that these resources are utilized inefficiently in 3D parallelism. 
For computation, synchronous PMP is widely accepted owing to its more stable convergence. But the GPU idle time, otherwise referred to as bubbles, decreases the throughput of PMP. 
For GPU memory, the activation recomputation technique is necessary when training giant models; however, existing 3D parallelism approaches~\cite{paddle,rasley2020deepspeed,megatronsc} simply apply this technique to all model layers, leaving parts of the GPU memory unutilized.
For both network bandwidth and computation, 
current TMP implementations~\cite{shoeybi2019megatron} need to communicate in each layer to obtain a complete output, which blocks the computations of subsequent layers; meanwhile, there is no data transmission during computing. This interdependence results in the inefficient utilization of both the bandwidth and computing resources.
The training of large models requires dozens to hundreds of devices, and each run costs tens of thousands of dollars. 
Therefore, it is necessary to employ any available device resources to accelerate training.
Accordingly, another concept behind Merak is \emph{the acceleration of training with improved integration of training resources.}

In this paper,
we present a 3D parallelism framework called \emph{Merak}, which orchestrates DP, PMP, and TMP, providing a user-friendly and efficient distributed foundation model training.
Merak features two novel modules to address the above issues. First, an \textbf{automatic model partitioner} enables the automatic partitioning, allocation, and training of models in a distributed setting, thereby facilitating the development of foundation models.
However, building such a distributed training scheme automatically is non-trivial, 
as the loading of complete model properties can easily surpass the capacity of a single device, and the presentations of community models are very distinct.
To overcome these challenges, we design a \emph{proxy node-based graph} for community models, which can obtain entire model graphs with significantly smaller overheads and realize auto TMP.
We also develop a \emph{graph-sharding algorithm} to shard various graphs into sequences of structured subgraphs that can be executed sequentially and appropriately form a PMP.
In addition, to relieve the burden on DNN developers, we present a parallelism method-agnostic API abstraction. The transparent API allows developers to reuse local training scripts with minimal modifications. 

The second module developed in Merak is an \textbf{efficient 3D parallel runtime engine} that integrates three techniques to improve efficiency. 
First, we design a \emph{shifted critical path pipeline schedule} for pipeline bubble reduction.
The critical path is an operation sequence that determines pipeline latency. Our schedule shortens the critical path by discarding redundant recomputations, as well as adjusting operation orders and start times.
We also observe that more efficient memory utilization can be achieved by means of fine-grained activation recomputation.
We therefore develop a \emph{stage-aware recomputation} method that employs idle worker memory to reduce recomputation according to pipeline stage rank and depth, thus speeding up training. 
Furthermore, we improve the concurrency of communication and computation in TMP using our \emph{sub-pipelined TMP} approach, which splits microbatches into individual sub-microbatches, and pipelines them to overlap the communication and computation.

Our contributions are summarized as follows:
\begin{itemize}
\item We present Merak, a distributed training framework for generalizing and accelerating 3D parallelism. 
We simplify 
the use of 3D parallelism, and generalize API to support community models.
To best of our knowledge, Merak is the first PyTorch-based framework that enables automatic 3D parallelization.
\item We develop a high-performance 3D parallel runtime engine to exploit available training resources. 
Our \emph{shifted critical path} pipeline schedule focuses on utilizing computational resources, \emph{stage-aware recomputation} concentrates on maximizing the use of memory resources, and \emph{sub-pipelined TMP} promotes the use of both computation and communication.
\item We demonstrate the efficiency of Merak by conducting comprehensive evaluations on models with up to 20 billion parameters at different training scales. Compared with baseline approaches including SOTA 3D parallelism library, Merak accelerates the training process with 1.18--1.61$\times$ throughput. 
We open the source code at \url{https://github.com/hpdl-group/Merak}.
\end{itemize}

\section{Background and Related Work}

A deep learning model is typically constructed using a sequence of layers and trained by multiple iterations. A typical DNN training step involves a forward pass (FP) and backward pass (BP). The iteration of steps refines the model to achieve higher accuracy. 
With the flexibility and expandability of transformer structures, transformer-based models~\cite{gpt2,t5model,mt-nlg,brown2020language,devlin2018bert} have continued to grow rapidly with renewed accuracy records. 
To reduce training time, 
training may be distributed among devices with particular parallel strategies at different scales. 
By combining data, pipeline model, and tensor model parallelisms, \textbf{3D parallelism}~\cite{megatronsc,ds3d,colossal,karakus2021amazon,whale} 
leverages their merits and becomes the SOTA distribute training method for giant foundation models. 

\textbf{Data parallelism (DP).} The most common way to accelerate the model training is DP~\cite{sergeev2018horovod,torchddp,bagua}, 
where data remains separate for different workers, while each worker holds a model replica and performs collective primitives, such as AllReduce~\cite{allreduce}, at certain intervals to synchronize the model. 
However, a single GPU cannot hold an extremely large model.
ZeRO~\cite{zero} of DeepSpeed~\cite{rasley2020deepspeed} eliminates memory redundancy by partitioning the model states among all workers. ZeRO-Offload~\cite{zero-offload}, ZeRO-Infinity~\cite{zeroinf}, and PatrickStar~\cite{patrickstar} further swap memory to the CPU and NVMe. They trade communication volumes to scale the model capacity of a single GPU.
However, as model size increases, the increase in communication limits training performance. DP generally requires few additional lines of code to single-GPU training scripts, thus offering an important usage template for Merak. All optimizations are orthogonal and can be adopted in the DP part of Merak.

\textbf{Pipeline model parallelism (PMP).} PMP splits a model into sub-modules and scatters them into a worker group. The batch data are also separated into microbatches and executed in a pipeline to improve device utilization and reduce communication. Asynchronous PMP approaches~\cite{pipedream, pipedream-2BW,hetpipe,ftpipe,kosson2021pipelined} take full advantage of pipelining by holding multiple model versions or asynchronously updating weights.
However, works~\cite{pipemare, varuna} 
demonstrate several issues associated with the convergence rates and final accuracy of asynchronous PMP methods.
GPipe~\cite{huang2019gpipe} first design a synchronous PMP schedule and integrate it into PyTorch~\cite{pipetorch}. DAPPLE~\cite{fan2021dapple} and Megatron-LM~\cite{megatronsc} employ modified pipeline schedules that reduce the usage of activation memory. Synchronous PMP suffers from bubbles,
Hippie~\cite{hippie} utilizes bubbles for communication with half-step-delayed parameter updating, and
Chimera~\cite{chimera} uses a bidirectional pipeline that reduces bubbles, but doubles the model memory. 
Merak enables a synchronous schedule with a shortened critical path in the PMP component.

\textbf{Tensor model parallelism (TMP).}  Unlike PMP partition model layers, TMP generally splits individual layers or operators across multiple workers. 
Megatron-LM~\cite{shoeybi2019megatron} analyzes the architecture of transformer-based models~\cite{transformer}, divides weight matrices along row or column dimensions, and adds AllReduce operations to ensure correctness. SageMaker~\cite{karakus2021amazon} enables a more efficient memory solution by means of Reduce-Scatter. A line of workers~\cite{2DTP,2.5DTP,3DTP} further expands TMP to include more dimensions of weight parameters and input tensors to reduce both the redundancy of activation and the communication overhead. 
Merak uses a similar TMP implementation to Megatron-LM, and other optimizations could be complementary to Merak. 
However, existing TMP methods bring a large number of blocked communication operations during both FP and BP, which greatly slows down the training phase. We therefore focus on improving the concurrency in TMP part of Merak.

\textbf{Activation recomputation.} Activations are intermediate outputs of FP and are consumed by BP for the gradient calculation. Activation recomputation~\cite{activationcheckpoint, dtr, Checkmate} techniques evict these values, and recompute them when required. 
Although this approach costs about $1/3$ more arithmetic overheads, it may conserve significant memory footprints and enable training with larger data microbatches, while preserving more model parameters on each device. Activation recomputation has been widely adopted~\cite{varuna,ftpipe,autosurvey}, particularly in PMP approaches, where workers may manage a large number of activations simultaneously. Merak employs and fine-tunes this method to improve memory efficiency.

\begin{figure}[bt]
    \centering
    \includegraphics[width=0.90\columnwidth]{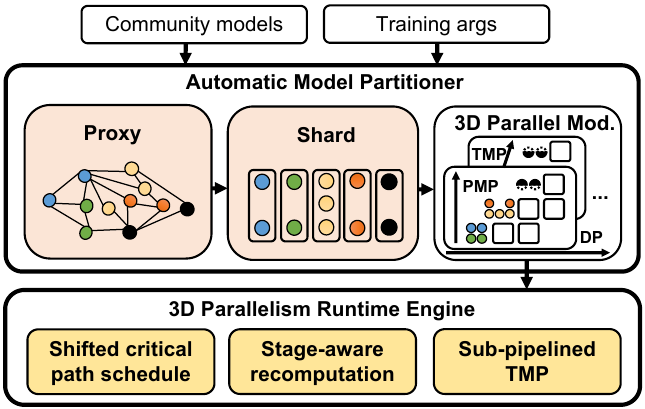}
    \caption{Merak overview.}
    \label{fig:overview}
\end{figure}

\section{Merak Overview}
We design Merak, a distributed training system based on the popular DNN framework PyTorch. A high-level overview of Merak is shown in Figure~\ref{fig:overview}. Merak features an automatic model partitioner and a 3D parallel runtime engine. 
The partitioner takes community models as input. As there is a wide range of model definition patterns, the input lacks a specific model layout.
To get perspectives of distinct models, we \emph{proxy} the models with proxy node-based graphs, allowing them to be handled with a single processor. 
Under proxied representation, we introduce a graph-sharding algorithm that \emph{shards} the model into proper subgraph slices, which can be executed sequentially and form pipelines with low communication volumes. 
According to the parallel settings, each worker obtains a part of the sequential subgraphs, and builds it into a 3D parallel module.
The model partitioner, which performs pre-processing before training, is described in further detail in Section~\ref{sec:autooverview}.

The 3D parallel runtime engine (Section~\ref{sec:runtime}) of Merak takes the results of the partitioner and training hyperparameters, such as global batch size. Our \emph{shifted critical path schedule} then arranges the execution schedule of training operations, including FP, BP, and recomputing, with a lower bubble ratio. In addition, our \emph{stage-aware recomputation} method reduces the activation recomputation overhead in terms of the relationship between pipeline depth and memory consumption. Furthermore, 
Merak conducts communication and computation in a non-blocking manner through \emph{sub-pipelined TMP} to improve the efficiency of TMP.
These Merak optimizations collaborate to significantly accelerate distributed training.

\section{Democratizing 3D Parallelism}  \label{sec:autooverview}
One key design principle of Merak is bridging the gap between complex 3D parallelism training and accessibility of community models.
We develop a model partitioner that creates a proxy representation of the model graph (Section~\ref{sec:proxy}) for auto TMP, and automatically shards the model into a subgraph sequence (Section~\ref{sec:shard}) for auto PMP. 
Subsequently, Merak assigns different subgraphs to workers according to a specific algorithm, and builds them into an execution module for the runtime engine.

With the help of our model partitioner, 
users do not need to account for the specific structure of each model, and are able to train giant models with 3D parallelism as easily as training on a single GPU.
An example script presented in Figure~\ref{fig:usage} shows that, compared to the standard training process using Transformers~\cite{Wolf_Transformers_State-of-the-Art_Natural_2020} Trainer, Merak only requires a few lines of code to set parallel degrees, and avoids model refactoring.
A very similar interface to that of Transformers Trainer offers convenience to training models in the Transformers library, where users can access a large number of model resources, as well as help from the active community.
Any model absent from the Transformer library can also be trained by Marek, so long as they are traceable. More examples, including those pertaining to various situations and tasks, are present in the Merak code repository.

\setminted{fontsize=\scriptsize}
\begin{figure}
\begin{minipage}{.46\linewidth}
\begin{minted}
[
% frame=leftline,
% fontsize=6pt,
]
{python}
import transformers as hf
...
...
...
# Set training args.
args=hf.TrainingArguments()
...
# Set trainer.
trainer=hf.Trainer(
    args=args,
    model=...,
    train_data=...,
    eval_data=...,
)
...
# Do train.
trainer.train()
\end{minted}
\end{minipage}
\vline \quad
\begin{minipage}{.48\linewidth}
\begin{minted}
[
% frame=leftline,
highlightlines={1,3,4,6,9},
% fontsize=6pt,
]
{python}
import Merak
# Init parallel degrees.
dp, tmp, pmp = 2, 2, 2
Merak.init(dp, tmp, pmp)
# Set our training args.
args=Merak.MerakArguments()
...
# Set our trainer.
trainer=Merak.MerakTrainer(
    args=args,
    model=...,
    train_data=...,
    eval_data=...,
)
...
# Do train.
trainer.train()
\end{minted}
\end{minipage}
\caption{Merak enables 3D parallelism with a few lines of code change. The left side shows the standard single GPU training script of Transformers~\cite{Wolf_Transformers_State-of-the-Art_Natural_2020} Trainer, whereas the right side shows the same model's training process through 3D parallelism with Merak.}
\label{fig:usage}
\end{figure}

\subsection{Proxied Graph and Automated TMP} \label{sec:proxy}
Since PyTorch offers a variety of model definition methods, we must obtain a uniform modality of disparate community model codes.
Merak employs the trace tool \textit{torch.fx}~\cite{torchfx} in PyTorch, which traces an entire model into a GraphModule, but it requires running the model for one step. However, training, even only loading, a DNN model can easily exceed the memory capacity of a single device nowadays. For example, a 175-billion-parameter GPT-3 model requires 700 GB of memory to store parameters.

To reduce the memory overhead,
we design \emph{proxy nodes} to patch intensive computing nodes, such as GEMM and convolution operations, which usually require the most computational load in recent DNN architectures.
Proxy nodes are not associated with any parameters, but they can return appropriately-sized results according to the inputs, and therefore participate in model tracing normally. Furthermore, proxy nodes remember all functionalities, and can be restored to the functional node after being assigned to GPUs.
The proxy node-based graph enables a single worker to store a whole giant DNN model and swiftly executes model tracing on CPUs, which typically have a larger memory capacity than GPUs. In our experiments, one regular server with 96GB of RAM could handle a complete GPT-3 model.

Furthermore, we implement the auto TMP with proxy node-based graph. As we proxy all GEMM operators and follow Megatron-LM~\cite{shoeybi2019megatron} for TMP implementation, proxy nodes will exactly include all operators that need to be partitioned. We use a feature dict to map a TMP attribute to each proxied node, which can provide information such as TMP degree, operator partition dimension, and communication primitive. 
With the help of TMP attributes, model tracing will correctly generate the TMP-enabled proxied model graph. When reconstructing the computation nodes from proxy nodes, we adopt TMP modules instead of the original GEMM operations and add communication operations according to TMP attributes, thus achieving automated TMP. We hold default feature dicts for common transformer-based models in community, including transformer decoder models (e.g. GPT~\cite{gpt2}), encoder models (e.g. BERT~\cite{devlin2018bert}),  encoder-decoder models (e.g. T5~\cite{t5model}), and image classification models (e.g., ViT~\cite{vit} and Swin Transformer~\cite{swin}). In addition, users can easily define a feature dict through our interface.

\subsection{Graph-Sharding Algorithm and Automated PMP} \label{sec:shard}

Although we have obtained traced graphs of community models, these graphs are still highly mutually distinct. 
To enable auto PMP, we must split the complete graph into a sequence of subgraphs, thus enabling us to construct a pipeline by assigning continuous parts of the subgraph sequence to each pipeline stage. 
Furthermore, because data transmissions between adjacent subgraphs are candidates for communication between stages, the number of nodes transferred between subgraphs should be minimal. 
Thus, the partitioning of distinct graphs into sequential subgraphs with limited connections is our fundamental problem for auto PMP. 
We propose a \emph{graph-sharding algorithm} that traverses all nodes, determines each node’s minimum dependency, and splits the graph into subgraphs according to the dependency at the lowest communication overhead between subgraphs.

The graph of DNN models is a directed acyclic graph (DAG), we can traverse all nodes with the order of DAG and create subgraphs from scratch.
To determine whether a node could stay in a subgraph, we find its farthest dependency and related subgraph id with a \emph{dependency search algorithm} as shown in Algorithm~\ref{alg:mindepend}. 
The farthest dependency of a node could be a graph input (lines 5-10), an output of previous subgraphs (lines 11-14), or other previously visited nodes (lines 15-19). 
We do not consider the last visited node as a potential dependency because the last visited node is either in the current subgraph, or the output of the last subgraph.
A node should stay in the same subgraph with its farthest dependency. Meanwhile, a node could be placed into a new subgraph if it has no dependency or its farthest dependency is the latest created subgraph. This procedure ensures only one node could be transmitted between adjacent subgraphs thus low communication volumes between stages.
However, the strict rules might yield no subgraphs at all, 
specially, we define a kind of node as \textit{common nodes} and allow them to become exceptions of dependency (lines 3-4) and pass them among subgraphs. 
Take transformer layer based model as an example, there are nodes asked by every transformer block, such as the attention mask node, thus we could refer nodes with a user count more than the number of transformer layers to the common nodes. 
We collect user counts of common nodes from graph tracing and inject the count into our shard algorithms. 
And during searching dependency, we update the user counts of common nodes to guide whether they should be returned by a subgraph, and record the first user of each graph input.

Next, we partition the graph with an auto \emph{graph-sharding algorithm} as shown in Algorithm~\ref{alg:graphshard}. 
We begin node traversing with subgraph id 0 and setting a subgraph id to each node with the help of farthest dependency results. 
Specifically, for a node and its farthest dependency, once the subgraph id of its farthest dependency is smaller than the current, we will update the current subgraph id, set the updated subgraph id to nodes that have a larger index than the farthest dependency node in the visited nodes list (lines 5-11), and assign subgraph id to the node being traversed (line 12).  
Moreover, to accelerate the shard procedure, we can provide an upper bound of partition number based on the model layer number or device number (lines 13-14). 
And if we create a new subgraph, we handle outputs of current subgraph and increase the subgraph id (lines 15-22).
After traversing all nodes, we create
subgraphs with corresponding nodes, inputs and outputs (lines 23-29). 
Inputs of each subgraph include all outputs of the last subgraph, thus the subgraph list could be executed as a sequence. 
Take the GPT series model as an example, each attention block and feed forward network block could become an individual subgraph module with our algorithms.

\begin{figure}[!t]
 \removelatexerror
\begin{algorithm}[H]
\small
\caption{Search Node Dependency}
\label{alg:mindepend}
\DontPrintSemicolon
\KwIn{node $n$, current subgraph id $s$, common nodes to user counts dict $\mathcal{C}$, graph inputs $X$, graph inputs to nodes dict $\mathcal{I}$, nodes to subgraph ids dict $\mathcal{M}$, subgraph ids to subgraph outputs dict $\mathcal{S}_{outs}$, visited nodes list $N$}
\KwOut{minimum subgraph id $s_{min} $, farthest dependency node $n_{min}$, updated $\mathcal{C}$ and $\mathcal{I}$} 
\BlankLine
$s_{min} \gets s, n_{min} \gets Null$ \\
\For{$arg \in n.args$}{
    \tcp{$arg$ is a common node, update its user count.}
    \uIf{$arg \in \mathcal{C}.keys$}{
        $\mathcal{C}[arg] \gets \mathcal{C}[arg] - 1$
    }
    \tcp{$arg$ is a graph input.}
    \uElseIf{$arg \in X$}{
        \uIf{$\mathcal{I}[arg] = Null$}{
             $\mathcal{I}[arg] \gets n$
        }
        \ElseIf{$\mathcal{M}[\mathcal{I}[arg]] < s_{min}$}{
            $s_{min} \gets \mathcal{M}[\mathcal{I}[arg]],\quad n_{min} \gets \mathcal{I}[arg]$
        }
    }
    \tcp{$arg$ is a output in previous subgraphs.}
    \uElseIf{$arg \in \mathcal{S}_{outs}.values$}{
        \If{$\mathcal{M}[arg] + 1 < s_{min}$}{
            $s_{min} \gets \mathcal{M}[arg] + 1,\quad n_{min} \gets arg $
        } 
    }
    \tcp{$arg$ is the output from other previous nodes but not from the last visited node.}
    \ElseIf{$arg \in \mathcal{M}.keys $ and $ arg \ne N[-1] $}{
        \If{$\mathcal{M}[arg] < s_{min}$}{
            $s_{min} \gets \mathcal{M}[arg],\quad n_{min} \gets arg $
        }
    }
}
\Return $s_{min}, n_{min}, \mathcal{C}, \mathcal{I}$
\end{algorithm}
\end{figure}

\begin{figure}[!t]
 \removelatexerror
\begin{algorithm}[H]
\small
\caption{Graph-Sharding}
\label{alg:graphshard}
\DontPrintSemicolon
\KwIn{traced graph module $G$, graph inputs $X$, number of subgraphs $k$, nodes to number of parameters dict $\mathcal{D}$, common nodes to user counts dict $\mathcal{C}$}
\KwOut{subgraph list $G_{list}$} 
\BlankLine
Init nodes to subgraph ids dict $\mathcal{M}$; graph inputs to nodes dict $\mathcal{I}$; subgraph ids to subgraph outputs dict $\mathcal{S}_{outs}$; visited nodes list $N$; \\
Init parameter count threshold $p \gets Sum(\mathcal{D}.values) / k$; current subgraph id $s \gets 0$ \\
\For{Node $n \in G$}{
    $s_{min}, n_{min}, \mathcal{C},\mathcal{I}  \gets SearchDependency(n, s, \mathcal{C}, X, \mathcal{I}, \mathcal{M}, \mathcal{S}_{outs}, N)$ \\
    \If{$s_{min} < s$}{
        $s \gets s_{min}$ \\
        \tcp{Previous nodes between $n$ and $n_{min}$ need stay in the same subgraph.}
        \For{$n_{prev}  \in  Reversed(N)$}{
            $ \mathcal{M}[n_{prev}]\gets s_{min} $ \\
            \lIf{$ n_{prev} = n_{min}$}{
                \textbf{break}
            }
        }
    }
    $ \mathcal{M}[n] \gets s, \quad N.append(n) $ \\
    \tcp{Get total parameter number in subgraph $s$.}
    $ p_s \gets Sum(\{ \mathcal{D}[n_i] \mid n_i \in G, \mathcal{M}[n_i] = s \})$ \\
    \If{$p_s > p$}{
        \For{$ n_c \in \mathcal{C}.keys$} {
            \tcp{If a common node is visited and has remaining users, it should be added to the output of this subgraph.}
            \If{$n_c \in \mathcal{M}.keys $ and $\mathcal{C}[n_c] > 0$}{
                $\mathcal{S}_{outs}[s].append(n_c)$
            }
        }
        $\mathcal{S}_{outs}[s].append(n)$, \quad $s\gets s+1$ \\
    }
}
\tcp{Create subgraphs with nodes, inputs and outputs.}
\For{$s_i \gets 0$ to $s-1$}{
    $G_{nodes} \gets \{ n_i \mid n_i \in G, \mathcal{M}[n_i] = s_i \} $ \\
    $G_{inputs} \gets \mathcal{S}_{outs}[s_i-1] + \{\mathcal{I}[n_i] \mid n_i \in G_{nodes}\} $ \\
    $G_{outputs} \gets \mathcal{S}_{outs}[s_i] $ \\
    $G_i \gets CreateGraph(G_{nodes}, G_{inputs}, G_{outputs})$ \\
    $G_{list}.append(G_i)$ \\
}
\Return $G_{list}$
\end{algorithm}
\end{figure}

The graph-sharding overheads can be ignored because the 3D parallel training process requires many days. 
However, it is common to tune training configurations and restart training procedures multiple times, and repeated graph-sharding operations incur considerable costs. 
Since the subgraph sequence results relate only to the model itself, rather than the training arguments, such as GPU numbers, we can cache the subgraph sequence to significantly reduce the overhead. Table~\ref{tab:algopef} lists the graph-sharding times for different model sizes.

\begin{table}[tb]
    \renewcommand{\arraystretch}{1.3}
    \centering
    \caption{A comparison of execution overheads (in minutes) for the graph-sharding algorithm on different model sizes.}
    \label{tab:algopef}
    \begin{tabular}{c c c}
    \hline
    Model & W/o Cache & Cache \\
    \hline
    GPT-1.4B & 6.25 & 0.19 \\
    \hline
    GPT-2.5B & 9.85 & 0.26 \\
    \hline
    GPT-175B & 48.57 & 0.48 \\
    \hline
    \end{tabular}
\end{table}

After getting the complete subgraph sequence result, we follow AutoPipe~\cite{liu2022autopipe} to assign different parts of the subgraph sequence to different devices and form pipeline stages, thus auto PMP is achieved. Take the GPT-2~\cite{gpt2} model training as an example, our model partitioner will split the model into 98 subgraphs (1 embedding subgraph, 48 attention subgraphs, 48 feed-forward network subgraphs, and 1 head layer subgraph), where up to two nodes being transmitted between adjacent subgraphs. When training with 4 pipeline stages, AutoPipe will assign 27, 26, 24, and 21 subgraphs to each stage to balance workloads between stages. Subsequently, each worker will restore the proxy nodes into functional modules, and construct communication groups for TMP, PMP, and DP. Now our high-performance distributed runtime engine can perform the training procedures.

\section{High-performance training} \label{sec:runtime}
Another design principle of Merak is the enhancement of model training performance. 
We propose a fast \emph{runtime engine} that utilizes more training resources with techniques including a shifted critical path schedule, stage-aware recomputation, and sub-pipelined TMP.
Merak benefits from these techniques simultaneously during model training tasks, which significantly improves training performance.

\begin{figure}[tb]
\centering
\subfloat[One forward one backward (1F1B) schedule, the most commonly used pipeline schedule in 3D parallelism approaches.]{
\includegraphics[width=0.98\linewidth]{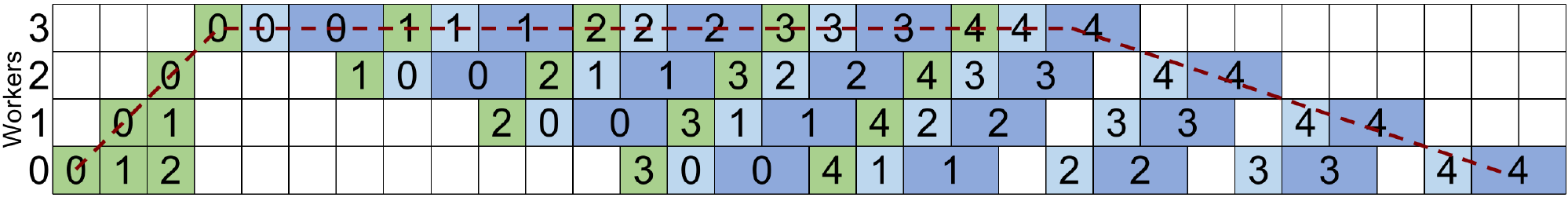}
}

\subfloat[Pipeline schedule when combining a 1F1B schedule with the early recomputation strategy of Gpipe \cite{huang2019gpipe}.]{
\includegraphics[width=0.98\linewidth]{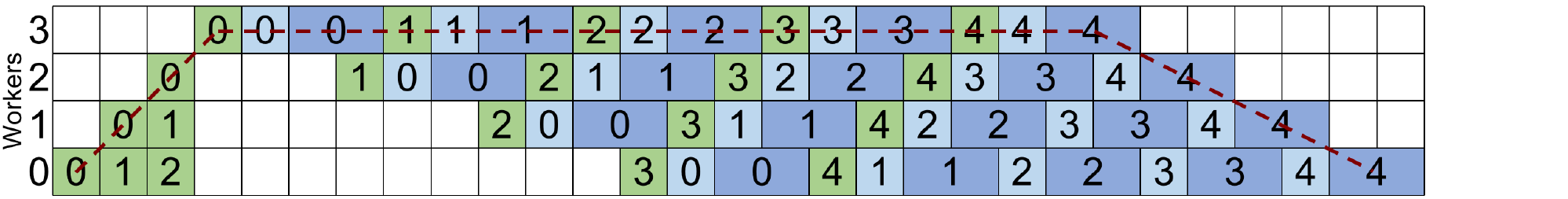}
}

\subfloat[Merak shifted critical path schedule, where the critical path is further shortened and shifted to Worker 2.]{
\includegraphics[width=0.98\linewidth]{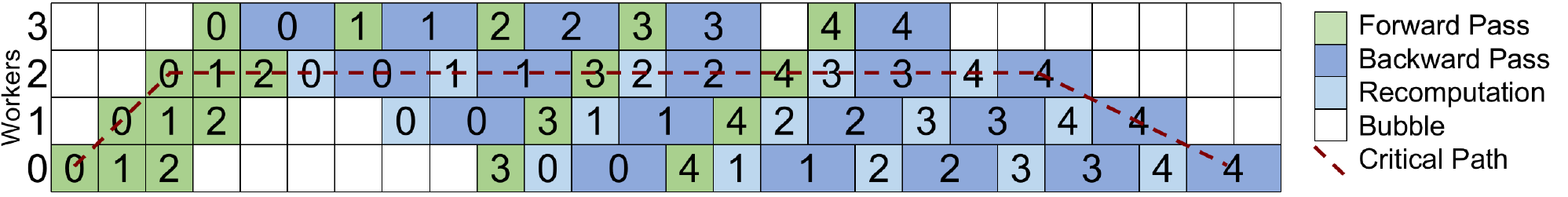}
}
\caption{Pipeline schedule comparison on four pipeline stages and five microbatches (0 to 4 in figures).}
\label{fig:pipetimeline}
\end{figure}

\subsection{Shifted Critical Path Schedule}
To ensure convergence and memory efficiency, Merak adopts synchronous PMP and activation recomputation with careful orchestration. 
We use the ratio of computation idle time (bubble time) to computation running time to measure the efficiency of the pipeline schedules. We denote the number of microbatches and pipeline stages by $m$ and $s$, respectively.
Assuming that the forward time of one microbatch is $T_m$, the recomputation and backward times can be estimated as $T_m$ and $2T_m$.
The most commonly used one-forward-one-backward (1F1B) schedule is shown in Figure~\ref{fig:pipetimeline}a, where the total bubble time is $(s-1)(T_m+T_m+2T_m) = 4(s-1)T_m$ and the running time is $4mT_m$. Thus, the bubble overhead ratio of the 1F1B schedule is $\frac{s-1}{m}$. This large bubble ratio greatly wastes the computational resources of workers.
The bubble time mainly lies in the start and end periods of pipeline schedules. Gpipe~\cite{huang2019gpipe} suggests that activation recomputation does not depend on the output of the previous stages. We apply this idea to the end period of the 1F1B schedule, as shown in Figure~\ref{fig:pipetimeline}b. With an earlier completed recomputation, we can reduce the bubble time to $(s-1)(T_m+2T_m) = 3(s-1)T_m$, and the bubble ratio becomes $\frac{3(s-1)}{4m}$.

\emph{Critical path} is a sequence of computation overheads across stages that determines the overall execution time of a pipeline. 
For the sake of simplicity, we assumed that the workloads are uniform among all stages.
As illustrated by the red dotted lines in Figure~\ref{fig:pipetimeline}, the critical paths of the 1F1B and early recomputation schedule are firmly pinned in the last stage. 
To shorten the critical path, we propose the \emph{shifted critical path schedule}, 
a pipeline schedule that shifts the critical path down by one pipeline stage. The schedule is based on the following observations: (a) when combining the 1F1B schedule and early recomputation of Gpipe, the recomputation in the final stage is redundant as the stage stores one activation, and (b) the critical path can be transferred to the penultimate pipeline stage when the final stage receives less computation. 
We drop the recomputation of the last stage, bring one FP computation of the penultimate stage ahead, fill it into the bubble, and adjust the recomputation in the first two BP processes accordingly. This optimization further reduces the bubble time to $3(s-2)T_m$, and the bubble ratio becomes $\frac{3(s-2)}{4m}$.

Furthermore, a typical transformer-based DNN model has head layers that project the outputs from transformer blocks to a given task, such as question answering and sequence classification. The head layers stay in the final stage and extend the critical path by damaging the balance of the pipeline stages, as the transformer blocks are distributed evenly.
But in the shifted critical path schedule, the computational workload of the final stage is lighter and can adequately handle extra calculations.
Consequently, the head layers exert little influence on the critical path.

\begin{table}[bt]
    \renewcommand{\arraystretch}{1.3}
    \centering
    \caption{Example memory (MB) and speed (samples/s) comparison on GPT-2.5B model and 8 RTX3090 GPUs (24 GB). The global batch size is 128 and the TMP degree is 1. MB size stands for microbatch size and Merak-SR stands for the Merak stage-aware recomputation methods.}
    \label{tab:dynamicactivation}
    \begin{tabular}{cccccc}
    \hline
    \# of  & MB & \multicolumn{2}{c}{Default } & \multicolumn{2}{c}{Merak-SR}  \\

    Stage & Size & Memory & Speed & Memory & Speed \\
    \hline
    \multirow{3}*{4} & 1 & 16720 & 6.81 & 21150 & 7.61 \\
     & 2 & 21262 & \textbf{6.84} & 21922 & 7.01 \\
     & 4 & OOM & - & OOM & - \\
    \hline
    \multirow{3}*{8} & 1 & 9066 & 6.39 & 21872 & \textbf{7.72}\\
     & 2 & 13130 & 6.64 & 19744 & 6.83\\
     & 4 & 21438 & 6.21 &  21818 & 6.39 \\
    \hline
    \end{tabular}
\end{table}

\subsection{Stage-Aware Recomputation} \label{sec:method2}

From the bubble ratio analysis, it is apparent that with a fixed global batch size and pipeline stage number, more microbatches will decrease the bubble ratio.
But a small microbatch size potentially affects the arithmetic intensity of operations. 
The default setting in Table~\ref{tab:dynamicactivation} lists several examples, which indicate that it is difficult to entirely run out of worker memory. There is a gap between the executable microbatch size 
and the out-of-memory (OOM) error when four stages are used, and the quickest configuration does not consume the maximal memory footprint with eight pipeline stages.
In other words, the remaining GPU memory resources are wasted, and can be used to further accelerate the process.

A straightforward solution is to reduce the use of activation recomputation to cut down the computation overhead. However, existing 3D parallelism libraries~\cite{paddle,rasley2020deepspeed,megatronsc} only provide open and closed options, and disabling recomputing leads to OOM errors.
Moreover, the memory overheads of non-checkpoint parameters vary among stages according to both pipeline depth and rank.
We perform the activation recomputation in a more fine-grained pattern by introducing \emph{stage-aware recomputation}, which is an activation recomputation scheme with dynamics among the pipeline stages.

The memory footprint of each stage relates to the number of parameters, microbatch size, and arrangement of PMP stages. Suppose that the runtime memory of each stage is $M_r$, which includes the activation of one microbatch $M_a$, model states, and temporary buffers. We can estimate $ M_r $ as a constant value between stages.
When $\alpha_i$ modules of the $i$-th stage do not use activation recomputation, 
they require an additional memory footprint $ (s-i) \alpha_i M_a$. Each stage should ensure that the total memory $M_r + (s-i)\alpha_i M_a $ does not exceed the device capacity. 
We expect the optimization goal to maximize the memory usage of all pipeline stages. To achieve this, we assume that in the optimal case each pipeline stage has the same memory consumption and each device is fully filled. This assumption is made to obtain the formula for the non-recomputation ratios between stages; that is, $M_r + (s - i) \alpha_i M_a = M_r + (s - j) \alpha_j M_a$ for stages $i$ and $j$. We then tune $\alpha_1$ by increasing it at intervals until an OOM error is caught. With the maximum $\alpha_1$, we calculate $\alpha_i$ for $i$ in $[2, s]$. But the non-recomputation ratios cannot larger than 1 in actual case, we set the smaller value between $\alpha_i$ and $1$ as the final $\alpha_i$ to ensure the correctness.
Since all pipeline stages use the maximum percentage of modules without activation recomputation, we achieve a high memory utilization.
Integrated with the shifted critical path schedule, we obtain the following recursive formula:
$$\alpha_i = \left\{ 
\begin{aligned}
    min(1, \frac{(s-1) \alpha_1}{s-i} ) & , &  i\in [2,s-1)\\
    \alpha_{s-2} & , &  i=s-1\\ 
    1 & , & i=s
\end{aligned}
\right.$$

The Merak-SR column in Table~\ref{tab:dynamicactivation} shows the impact of stage-aware recomputation. Compared with the default setting, Merak takes advantage of idle GPU memory, thus achieving achieve a 1.02--1.21$\times$ speedup. We can only compare the fastest configurations, and stage-aware recomputation reaches 1.13$\times$ higher throughput.

\begin{figure}[tb]
\centering
\subfloat[Default TMP. The interdependence between communication and computation leads to stalling in both streams.]{
    \includegraphics[width=0.97\linewidth]{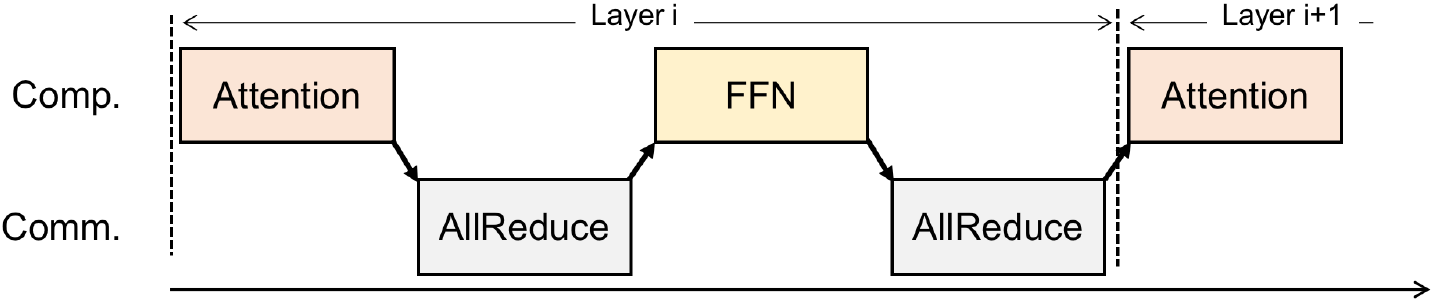}
}

\subfloat[Sub-pipeline TMP of Merak. The microbatch splitting enables pipelining operations. Computation and communication can be executed simultaneously.]{
    \includegraphics[width=0.97\linewidth]{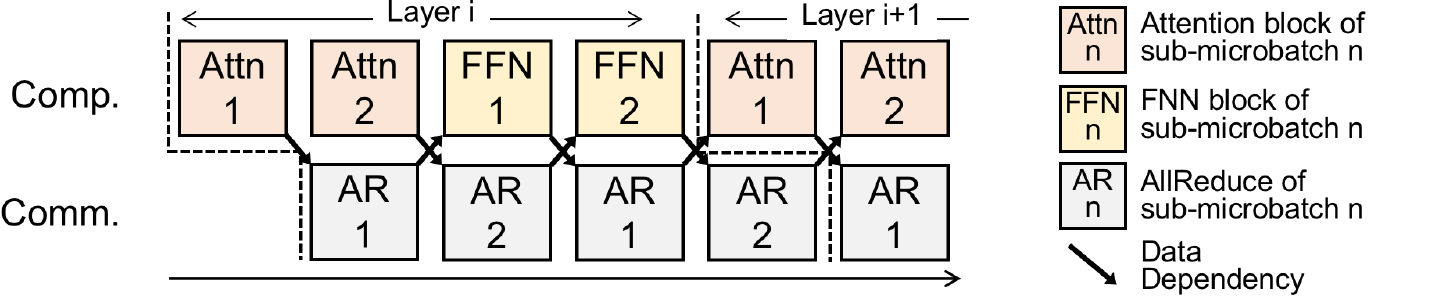}
}
\caption{Execution timelines of computation (Comp.) and communication (Comm.) stream on different methods.}
\label{fig:pipedtp}
\end{figure}

\begin{figure}[tb]
\centering
\includegraphics[width=0.9\linewidth]{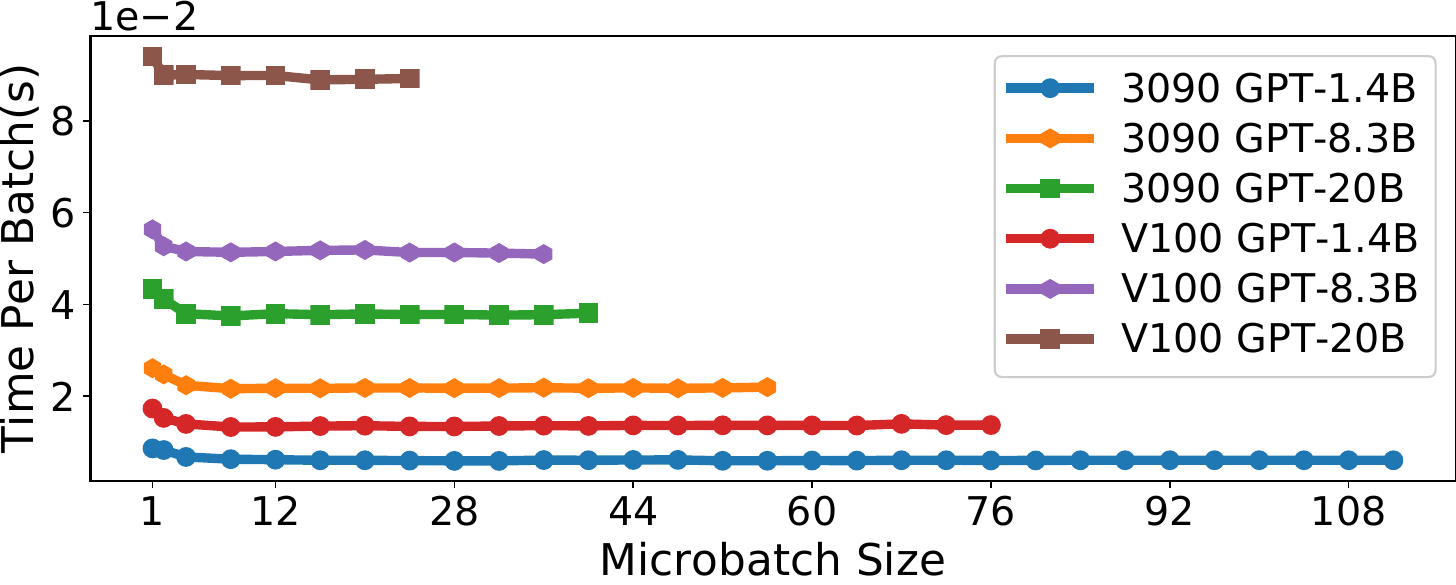}
\caption{Per microbatch size running time of a single transformer layer for different model sizes on RTX3090 and V100 GPUs.}
\label{fig:batchspeed}
\end{figure}
\subsection{Sub-pipelined TMP}

When training a giant model, sometimes a single worker cannot hold even a single transformer layer. 
For example, one layer of GPT-3 consumes over 34GB of runtime memory with a hidden size of 12288, sequence length of 2048 and microbatch size of 2. 
TMP~\cite{shoeybi2019megatron} is a solution that reduces the memory consumption of each device almost linearly.
A regular transformer layer can be divided into an attention block and a feed forward network (FFN) block~\cite{transformer}. Figure~\ref{fig:pipedtp}(a) presents a sample FP execution timeline for transformer-based model training with TMP.
TMP brings two AllReduce communications during FP, as well as the two AllReduces in BP, share a similar scheme.
All operations in TMP are exclusive, which is detrimental to training performance.

Pipelining is a common optimization design for stall reduction, and most DNN frameworks support the simultaneous execution of communication and computation streams. To fully utilize both streams, 
we proposed the \emph{sub-pipelined TMP} approach, which eliminates the dependency between computation and communication by pipelining microbatches, thus improving the efficiency through overlap. 
First, we measure the performance per sample of individual transformer layers from different models on two types of GPUs, along with increasing microbatch sizes.
The results are shown in Figure~\ref{fig:batchspeed}. The computation efficiency of workers becomes stable when the microbatch size is greater than 2; under the microbatch sizes of 1 and 2, the performance loss is less than 20\%, and tends to be smaller for larger models.
We also evaluate the efficiency of AllReduce operations; the performance change is narrow because the communication volume is sufficient, even under a microbatch size of 1. 
These observations indicate that we can scatter the microbatches and train them sequentially at a negotiable cost. 

\begin{figure*}[tb]
    \centering 
    \subfloat[GPT-1.4B, 1.19--1.42$\times$]{
        \includegraphics[width=0.485\linewidth]{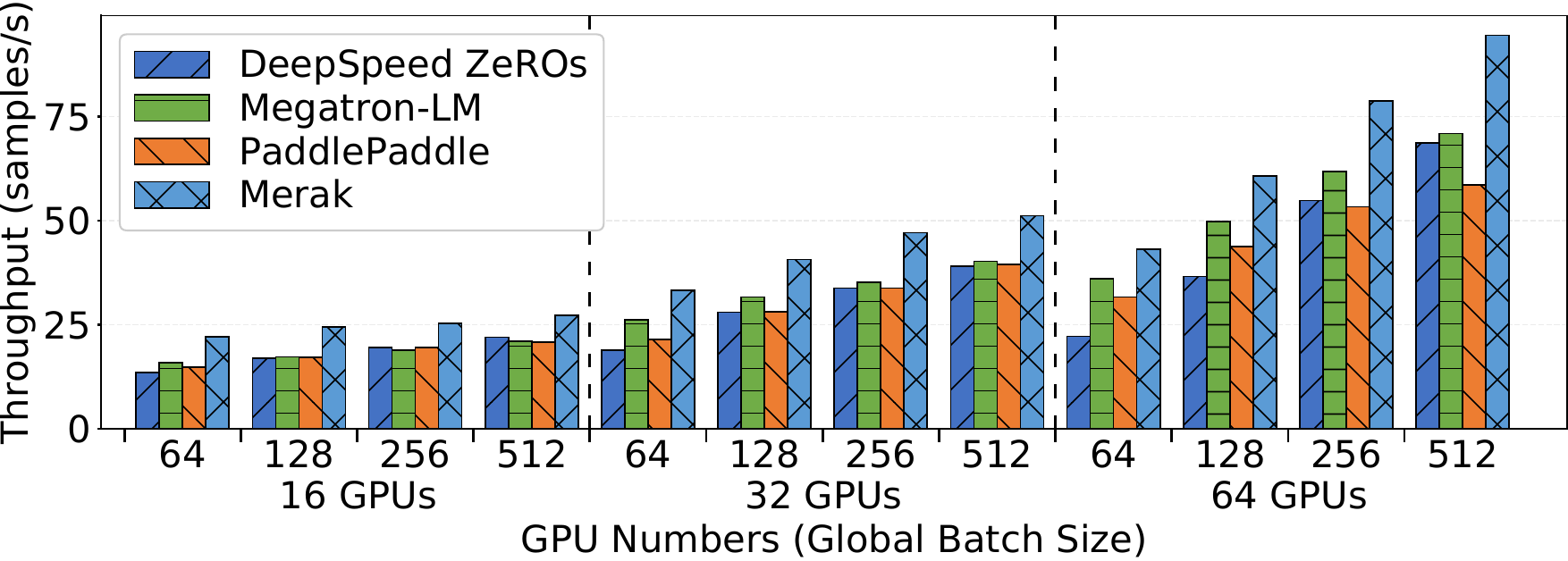}
    }
    \subfloat[GPT-2.5B, 1.19--1.39$\times$]{
        \includegraphics[width=0.485\linewidth]{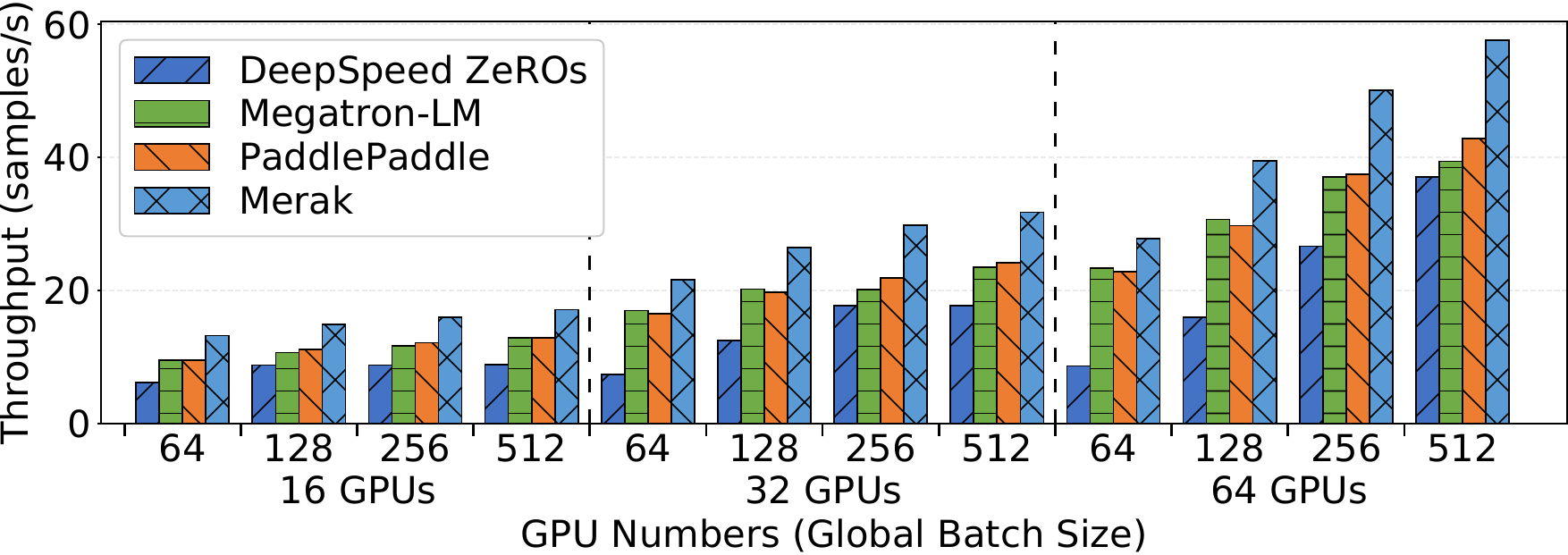}
    }
    
    \subfloat[GPT-8.3B, 1.18--1.43$\times$]{
        \includegraphics[width=0.485\linewidth]{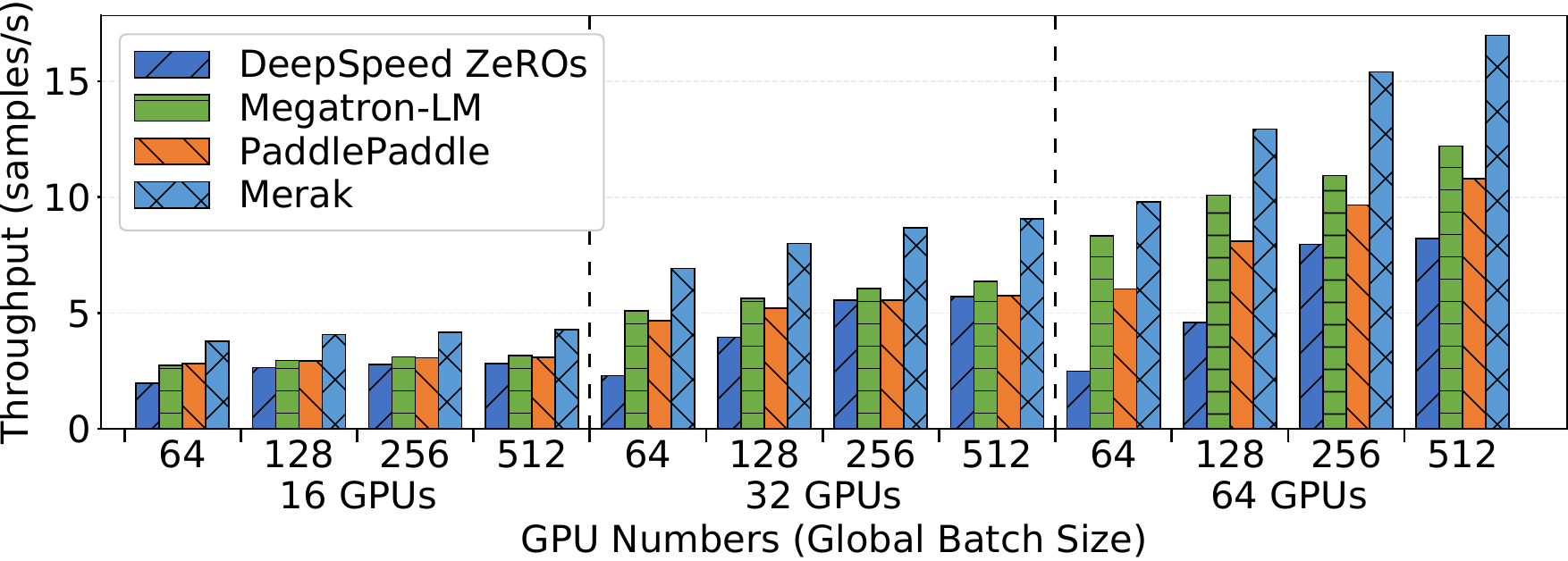}
    }
    \subfloat[GPT-20B, 1.50--1.61$\times$]{
        \includegraphics[width=0.485\linewidth]{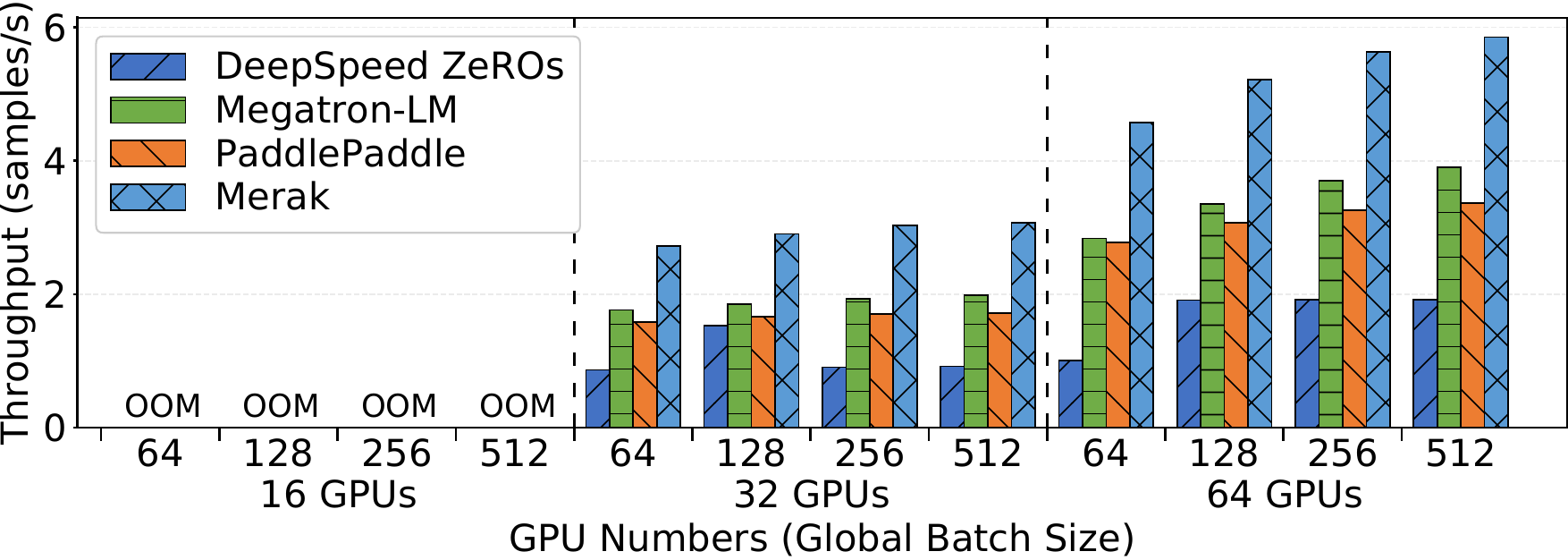}
    }
    \caption{End-to-end training throughputs on different number of GPUs and global batch sizes on four model sizes. The numbers in the captions denote speedups of Merak over the best baseline.}
    \label{fig:thourghput}
\end{figure*}

In the sub-pipelined TMP, we evenly split each microbatch of TMP blocks into two sub-microbatches, whose training procedures are mutually independent. We construct an inner pipeline with sub-microbatches: while one sub-microbatch communicates, the other sub-microbatch performs calculations, and vice versa. 
Figure~\ref{fig:pipedtp}(b) illustrates a sample FP timeline, where the communication and computation of sub-microbatches overlap across transformer layers, thus ensuring efficient usage of network bandwidth and computation resources. 
Let $T_m$ and $T_a$ denote the computational and communication overheads of one transformer layer during the FP. The overheads during the BP can be represented as $2T_m$ and $T_a$. Therefore, in a TMP module with $K$ transformer layers, one microbatch will cost a total of $(3T_m+2T_a)K$ with the default TMP approach. 
For simplification, we assume that the attention and FFN blocks have similar loads.
In sub-pipelined TMP, the cost of a microbatch becomes $ \frac{1}{4}T_m + (K-\frac{1}{4}) \cdot max\{T_m, T_a\} + \frac{1}{4}T_a $ in the FP and $ \frac{1}{2}T_m  + (K-\frac{1}{4}) \cdot max\{2T_m, T_a\}+ \frac{1}{4} T_a$ in the BP. And the total overhead is reduced to $\frac{3}{4}T_m + \frac{1}{2}T_a + (K-\frac{1}{4}) \cdot max\{3T_m, 2T_m+T_a, 2T_a\}$. We further report on method efficiency in Section~\ref{sec:evalmethod3}.

\section{Evaluation} \label{sec:eval}

\subsection{Experimental Setups}
\textbf{Platform Configurations.}
All experiments are conducted on a high-performance cluster with 16 server nodes. Nodes are connected via a 100Gbps InfiniBand.
The hardware of each node includes four NVIDIA GeForce RTX3090 GPUs each with 24GB memory, two Intel Xeon 4214R@2.40GHz CPUs, and 96GB DDR4 RAM. Intra-node GPUs interconnect via 16-lane PCIe 3.0.
All servers are run 64-bit Ubuntu 18.04, CUDA 11.3, cuDNN 8.2.1, NCCL 2.10.3~\cite{nccl}, and PyTorch 1.10.0. 
As the GPUs in each node run simultaneously, we only present the total number of GPUs used in the following experiments.

\textbf{Models and Datasets.}
We evaluate Merak’s performance using GPT-like transformer-based models. We follow the parameter settings used in prior studies~\cite{shoeybi2019megatron,varuna}, and adjust the hidden dimension and transformer layer numbers for different model sizes. 
We use four models with different sizes: GPT-1.4B, GPT-2.5B, GPT-8.3B, and GPT-20B, where B denotes the number of parameters in billions. The hidden size configurations are 1536, 1920, 3072, and 4096, respectively.
The benchmark models are trained with the OpenWebText~\cite{openwebtext} dataset, with a fixed sequence length of 1024 for all cases. Merak is a synchronous method that does not compromise convergence accuracy; therefore, we focus on comparing training performance. All reported values are averaged over 100 training steps.

It is worth noting that although we only evaluate one architecture, Merak can adapt to varying tasks and workloads effortlessly due to its user-friendly design. Since many popular models such as BERT~\cite{devlin2018bert} and ViT~\cite{vit} are based on transformer layers, we select the transformer decoder-based GPT model for its representativeness in large-scale pretraining.

\textbf{Baselines.}
Three approaches are compared with Merak in our experiments: (i) \textbf{DeepSpeed ZeROs}~\cite{rasley2020deepspeed,zeroinf,zero}. ZeRO technologies supported by our cluster including ZeRO-1, ZeRO-2, ZeRO-3, and ZeRO-Infinity. These strategies for trade-offs between device capacity and communication volume greatly affect training speed. Likewise, different model sizes, device numbers, and microbatch sizes limit the choice of strategy. 
Therefore, comparing with any single ZeRO approaches is not fair enough, we conduct experiments on all ZeRO methods to present data from the configuration that achieved the best performance. 
(ii) \textbf{Megatron-LM}~\cite{shoeybi2019megatron, megatronsc}.
Starting from the hybrid parallelism of DP and TMP, Megatron-LM integrates PMP recently. Its 3D parallelism is the SOTA implementation for large-model pretraining. 
To ensure a fair comparison, we grid-search the space of setting (TMP degree, PMP degree, and microbatch numbers) in Megatron-LM to find the best performance for each training task. We will only compare the best performance unless otherwise specified.
(iii) \textbf{PaddlePaddle}~\cite{paddle}. Unlike DeepSpeed and Megatron-LM, PaddlePaddle is an individual DNN framework that integrates 3D parallelism, but also adopts model state sharding between DP groups, similar to ZeRO-1. 
As with all other baselines, we tune the performance of PaddlePaddle for every experimental task and present the best results. 

Megatron-LM and PaddlePaddle require specific formatted models to enable 3D parallelism; we therefore use implementations from official examples.
Merak and DeepSpeed ZeROs use community implementations of Transformers 4.15~\cite{Wolf_Transformers_State-of-the-Art_Natural_2020} and do not need any changes in model codes.

\subsection{End-to-End Training Performance}

Figure~\ref{fig:thourghput} shows a performance comparison between the four systems for different model and training resource settings. Owing to our high-performance runtime engine, 
Merak is the fastest system in all cases, and achieves 1.18--1.61$\times$ speedups over the best baseline. 
We discuss the experiment detail of each model as follows.

\textbf{GPT-1.4B.} 
With the smallest model size, Merak can only enable the shifted critical path schedule and stage-aware recomputation because the TMP degree is 1 for the best configurations of all 3D parallelism methods. 
The relatively low memory overhead of GPT-1.4B provides substantial space for stage-aware recomputation, whereas it is insufficient for turnoff activation recomputation in other methods. 
It is worth mentioning that on 16 GPUs with global batch sizes (GBS) of 256 and 512, DeepSpeed ZeROs is the fastest baseline owing to the accessibility of ZeRO-1 and fewer parameter updating operations. ZeROs fall behind in all other experiments due to their small executable microbatch size, as well as large communication volumes in ZeRO-3 and ZeRO-Infinity.
For GBS of 64 and 128 on 64 GPUs, Merak obtain an acceleration of only 1.19--1.22$\times$, as high DP degree (8--16) and small GBS yield a small number of microbatches, and the DP communication and model updates consume a considerable portion of the runtime.
Merak demonstrate high performance in other scenarios, with speedups of up to 1.42$\times$. 

\textbf{GPT-2.5B.}
The TMP degree remains 1 for GPT-2.5B, which produce a similar performance trend to that of GPT-1.4B.
Merak outperforms the best baseline by 1.32--1.39$\times$, 1.27--1.36$\times$, and 1.19--1.34$\times$ on 16, 32, and 64 GPUs, respectively.
With different operator implementations, PaddlePaddle has an edge on small number of GPUs, whereas Megatron-LM performs better on larger scales. But with the help of ZeRO-1-like model sharding, PaddlePaddle performs well in larger GBS on 32 and 64 GPUs with GPT-2.5B, where sharding can conserve memory for a larger microbatch size.
In other cases, memory conservation alone is insufficient for extending the microbatch sizes.
As the model-sharding technique is orthogonal to Merak, we will attempt to integrate it in future studies to increase training flexibility.

\begin{figure}[tb]
    \centering
    \includegraphics[width=0.95\linewidth]{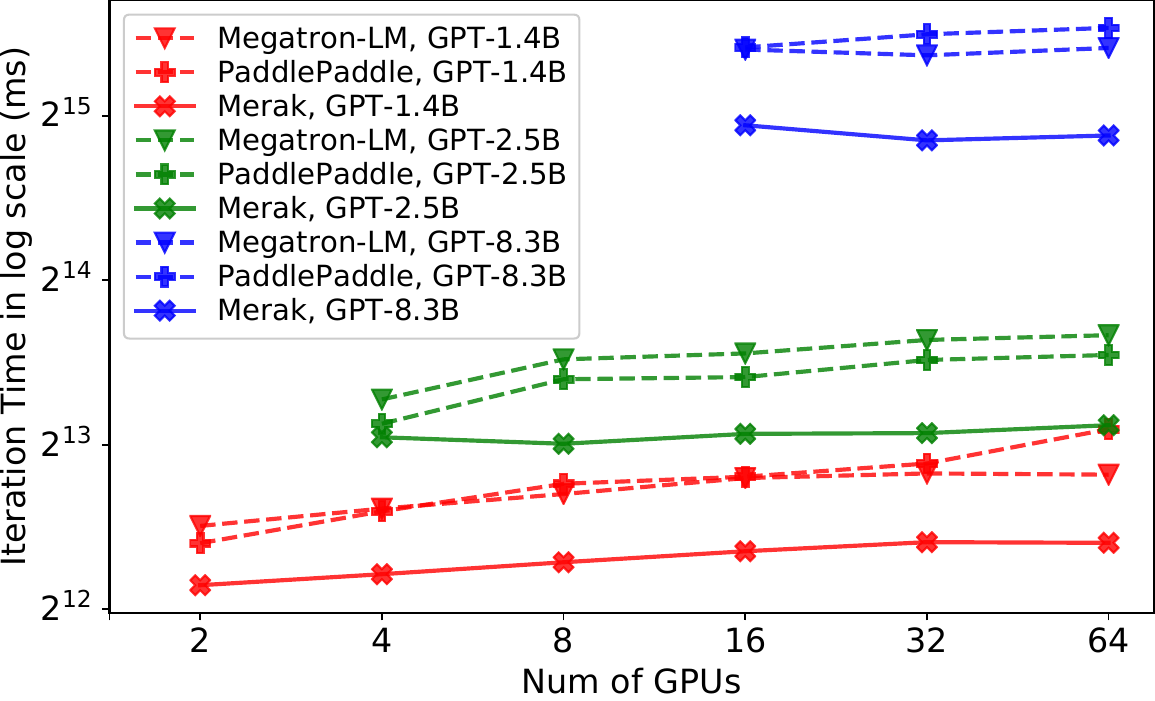}
    \caption{Weak scaling comparison on different model sizes. As the GPU number scales from 2 to 64, the global batch size scales from 16 to 512. The iteration time is presented in logarithmic scale and the missing data indicates OOM.}
    \label{fig:scalability}
\end{figure}

\begin{figure}[tb]
    \centering
    \includegraphics[width=0.95\linewidth]{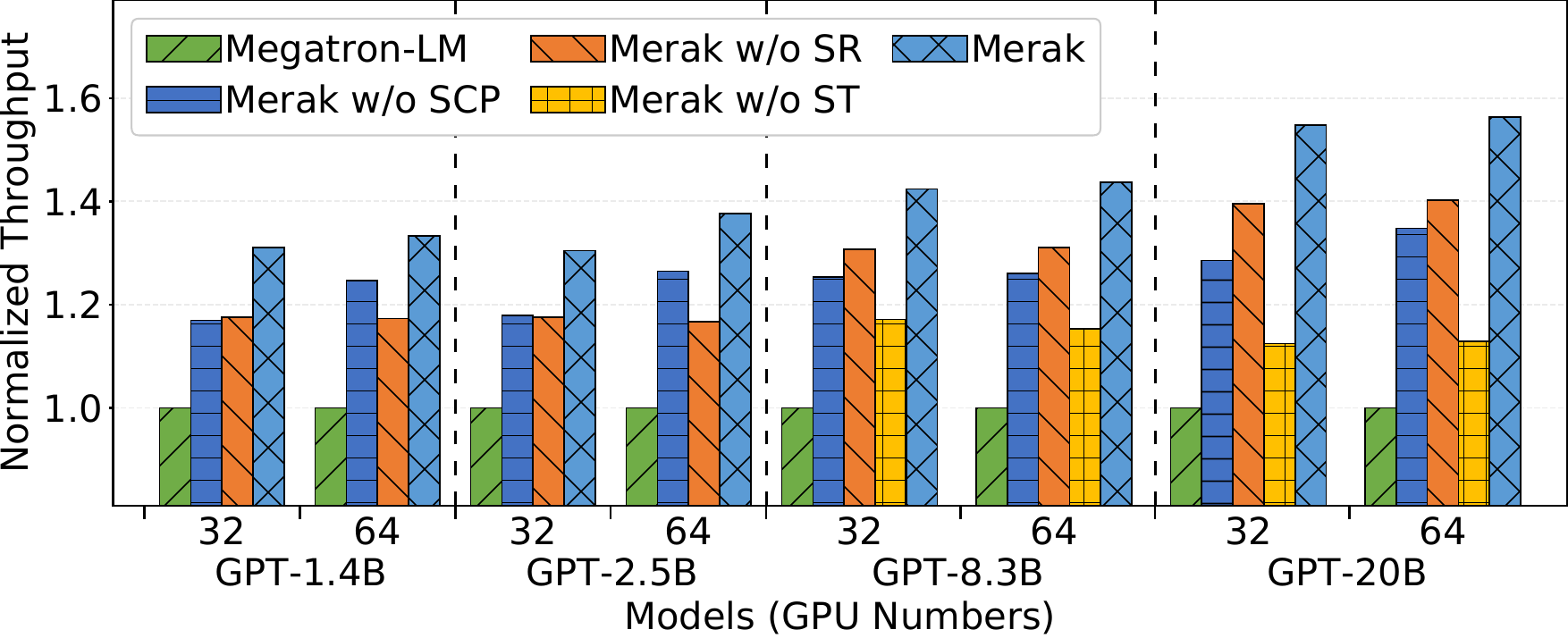}
    \caption{Training performance of Merak, Merak without shifted critical path schedule (SCP), Merak without stage-aware recomputation (SR), and Merak without sub-pipelined TMP (ST). Throughputs are normalized by Megatron-LM.}
    \label{fig:ablation_all}
\end{figure}

\begin{figure*}[bt]
    \centering
    \subfloat[Iteration time comparison between Merak shifted critical path (SCP) and 1F1B pipeline schedule for different PMP degrees. Lower iteration time indicates higher performance.]{
    \includegraphics[width=0.31\linewidth]{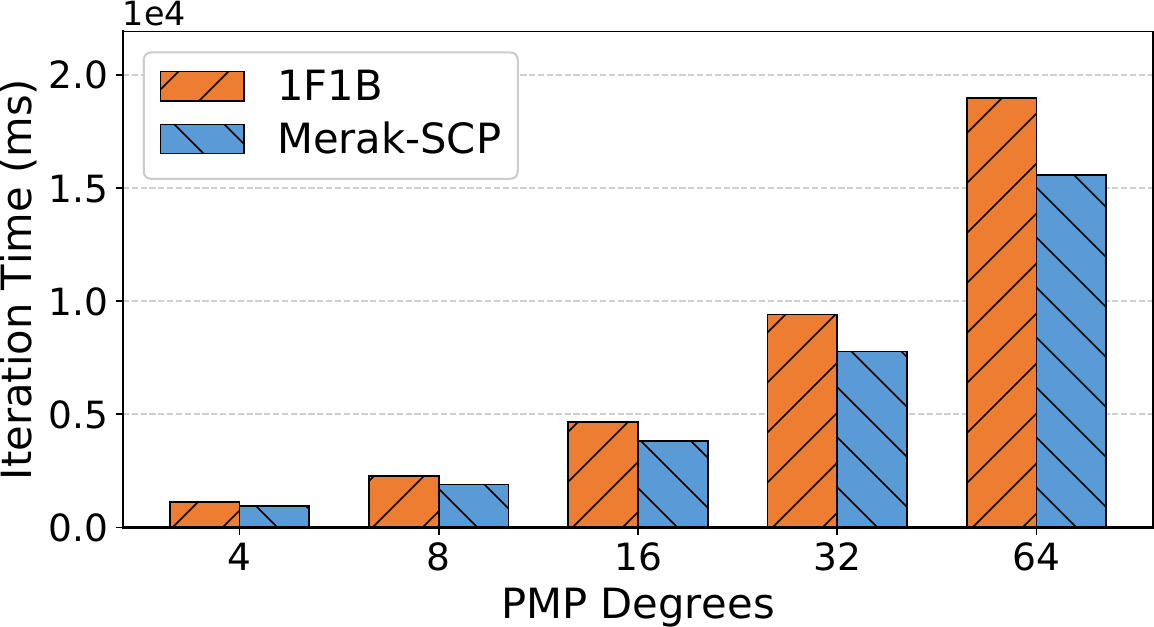}
    \label{fig:ablation_schedule}
    }
    \hfill
    \subfloat[Changes in training throughput and average memory consumption for GPT-2.5B on eight GPUs when tuning the non-recomputation ratio of stage-aware recomputation.]{
    \includegraphics[width=0.31\linewidth]{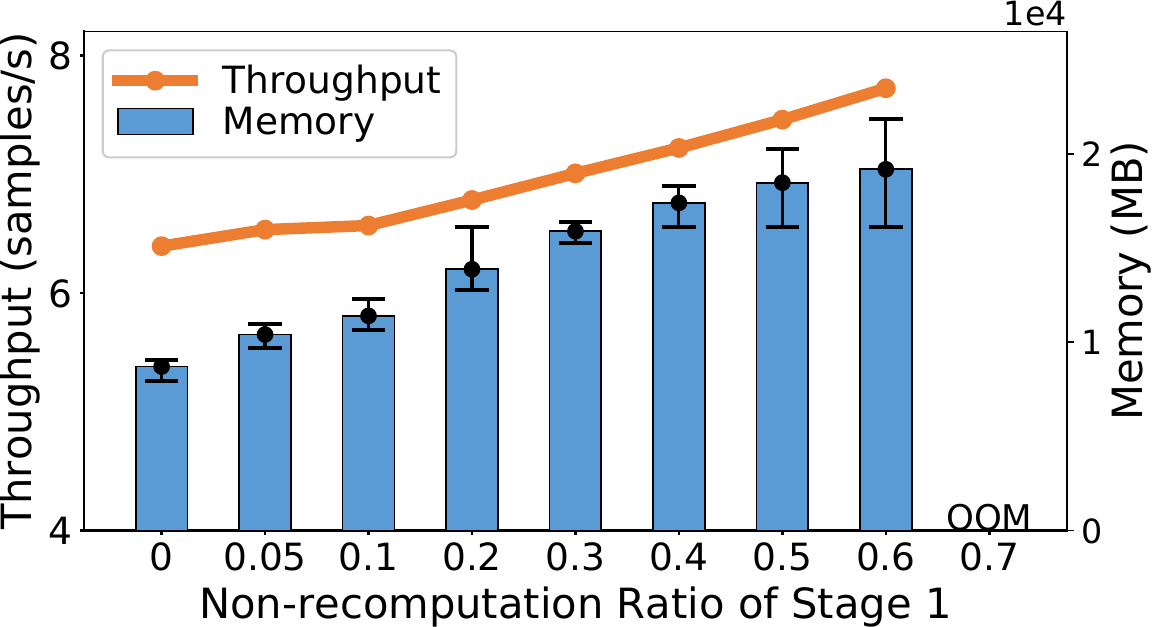}
    \label{fig:ablation_ac}
    }
    \hfill
    \subfloat[Iteration time comparison between Merak sub-pipelined TMP (ST) and Megatron-LM TMP for different TMP degrees. Lower iteration time indicates higher performance.]{
    \includegraphics[width=0.31\linewidth]{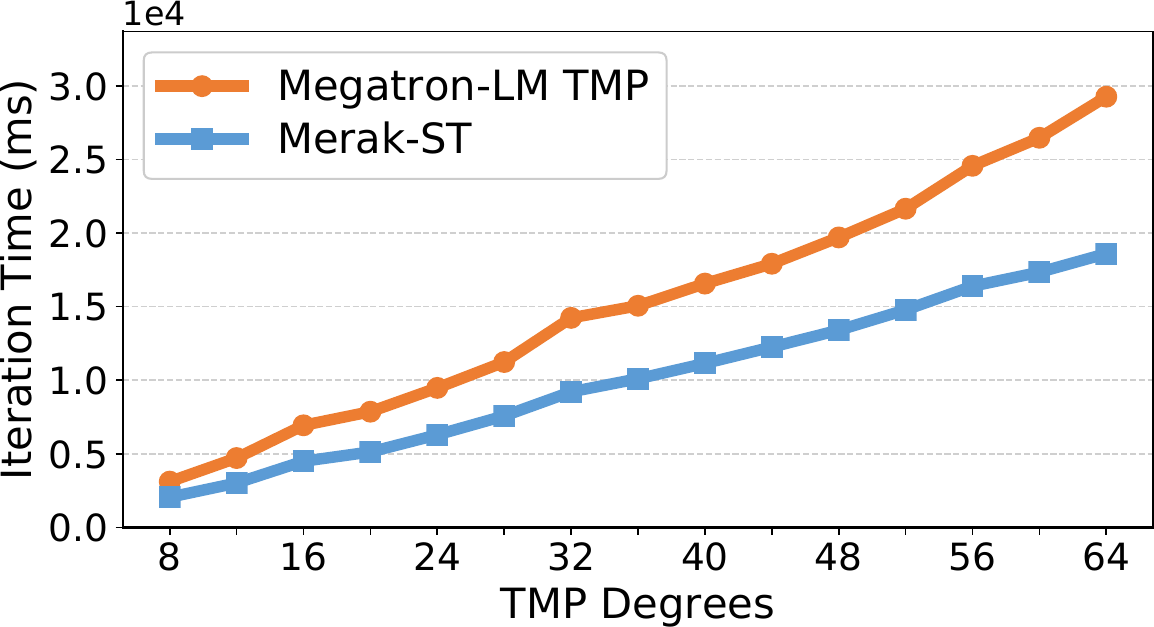}
    \label{fig:ablation_pipetp}
    }
    \caption{Impact of Merak optimizations.}
    \label{fig:ablation_mix}
\end{figure*}

\textbf{GPT-8.3B.}
When the model size expands to 8.3B, TMP is necessary to enable 3D parallelism methods. 
Although the speedups from the stage-aware recomputation become minor due to the larger model states,
Merak benefits from sub-pipeline TMP, and achieves an considerable acceleration of at least 1.34$\times$ in most cases. 
The exceptions of GBS 64 and 128 on 64 GPUs differ from relatively small models:
the PMP degree increases, whereas our sub-pipeline TMP requires a microbatch size of at least 2, which results in a decrease in the number of microbatches and increase in pipeline bubble ratios compared to Megatron-LM, whose microbatch size is 1. 
Our performance gains are maintained at 1.18--1.28$\times$ owing to the sub-pipeline TMP. But when GBS increases on 64 GPUs, the best microbatch size of Megatron-LM becomes 2, and the benefits of the shifted critical path schedule return. Now Merak’s speedups tend to be more stable, at a minimum of 1.39$\times$. Merak performs well on other scales, with speedups ranging from 1.34 to 1.43$\times$.

\textbf{GPT-20B.} 
For the largest model used in our experiments, no method can run successfully on 16 GPUs. It is worth noting that for 128 GBS and 32 GPUs, the maximum available microbatch size of DeepSpeed ZeRO-Infinity is 4. Although this yield a relatively satisfactory performance, the microbatch size drops back to 2 for larger GBS because gradient accumulation requires extra memory.
The communication of ZeRO-Infinity significantly limits training performance, particularly in large models. In Merak, however, the further TMP usage provides more potential improvement, and the number of microbatches becomes sufficient under lower degrees of PMP and DP. 
As our design can function fully, Merak demonstrate stable and significant acceleration compared to the fastest baseline, with 1.54--1.57$\times$ speedups on 32 GPUs, and 1.50--1.61$\times$ speedups on 64 GPUs.

\subsection{Scalability}

To demonstrate the scalability of Merak, we compare the iteration times of 3D parallelism approaches on benchmark models. For each model size, we scale the GPU numbers along with GBS. We present the shortest iteration time after searching for the configuration of parallel degrees and microbatch sizes for all baselines. The weak scaling results are shown in Figure~\ref{fig:scalability}. 

In the GPT-1.4B model, the high proportion of communication slows training, especially under inter-machine communication, and all methods display a obvious performance drop.
In the GPT-2.5B and 8.3B models, Merak exceeds linear scaling in certain cases. This is because when scaling the GPU number, Merak increases PMP degree instead of DP, enabling it to benefit more from stage-aware recomputation. 
In other situations, all 3D parallelism methods exhibit good scalability and similar performance trends. Consequently, Merak achieves stable speedups at most training scales.
We omit the results for GPT-20B because it requires at least 32 GPUs, and all methods scale well; for example, the speed loss of Merak is only 3.3\% when GPUs scaled from 32 to 64. 
Due to the limitations of our cluster scope, we cannot conduct tests on more GPUs. 
With the help of good scalability, we expect Merak to demonstrate continuous advantages in larger-scale training.

\subsection{Effectiveness of Optimizations}
We now demonstrate the contributions of Merak's optimization to end-to-end training performance. 
We conduct an ablation study on different model sizes and GPU numbers, and the results are shown in Figure~\ref{fig:ablation_all}. To effectively compare the benefits of each optimization, we fix the GBS to 512 and ensure that all methods use the same parallel degrees and microbatch sizes in each test.

All three optimizations are crucial in accelerating training, each of them provides different strengths. 
Owing to the same number of microbatches, the shifted critical path schedule provides relatively stable benefits in all cases.
Stage-aware recomputation plays a larger role in the relatively small models, GPT-1.4B and GPT-2.5B, because of their more idle memory. 
Stage-aware recomputation performs better on 64 GPUs because the best configuration of Merak scales the PMP degrees on small models, and generates a larger GPU memory optimization space.
Our sub-pipelined TMP provides significant in large model training where TMP is unavoidable. 
In the case of GPT-20B, which greatly suffers from blocked AllReduce operations in TMP, Merak results except Merak without sub-pipelined TMP exhibits significant advances in performance.
We further discuss the impact of these techniques in the following section.

\subsubsection{Shifted critical path schedule}

Figure~\ref{fig:ablation_schedule} shows the iteration time with the Merak shifted critical path schedule and commonly used 1F1B pipeline schedule. To ensure fair comparisons, we disable other Merak optimizations, and implement the 1F1B pipeline schedule based on the Megatron-LM open-source code.
To show the differences in pipeline schedules, we set the number of GPUs to PMP degrees, the number of microbatches to twice the PMP degrees, and assign two GPT transformer layers with a hidden size of 3072 to each GPU for the same workload. We scale the PMP degrees from 4 to 64, and the overall model size from 0.9B to 14.4B. 
Our schedule presents two major advantages: a lower bubble ratio from the shortened pipeline critical path, and a reduced pipeline imbalance caused by the computation of head layers in the final stage. 
Consequently, our shifted critical path schedule can reduce the per-step execution time by 18.6--22.0\%, and the time reduction is maintained for all GPU scales.

\subsubsection{Impact of stage-aware recomputation}
To verify that Merak achieves better memory efficiency with stage-aware recomputation, we train the GPT-2.5B model with PMP degree 8 and GBS 128 on eight GPUs. 
Figure~\ref{fig:ablation_ac} presents the throughput along with corresponding average memory usage among the pipeline stages for different non-recomputation ratios of the first stage ($\alpha_1$ in Section~\ref{sec:method2}). 
Training performance and memory utilization both increase with $\alpha_1$. 
For the non-recomputation ratios larger than 0.4, the minimal memory usage over stages stops increasing because the non-recomputation ratios of the latter stages reach 1, whereas the memory usage of the front pipeline stages continues to increase and so as the training speed. 
Finally, we detect an OOM error when 70\% of the layers of stage 1 do not use activation recomputation. 
Compared to disabling stage-aware recomputation ($\alpha_1 = 0 $), the training throughput can be improved by 1.21$\times$.

\subsubsection{Sub-pipelined TMP} \label{sec:evalmethod3} 
A iteration time comparison between the Merak sub-pipelined and Megaton-LM TMPs are illustrated in Figure~\ref{fig:ablation_pipetp}. 
To ensure an unbiased evaluation, we integrate Megatron-LM TMP in Merak, fix the GBS and microbatch size, disable stage-aware recomputation, employ our shifted critical path schedule, and train a 24-layers transformer-based model.
We set the TMP degree to the number of GPUs, and scale the TMP degrees from 8 to 64. 
The hidden dimensions of the models are also scaled from 1024 to 8192 so as equally divided parameter sizes. 
With overlapped communication and computation operations, sub-pipelined TMP improves the device efficiency and boosts the training in all situations, achieving 1.46--1.57$\times$ speedups over the Megatron-LM TMP. These consistent performance gains of sub-pipelined TMP comes from the significantly reduced idle time of both communication and computation streams. It is worth noting that both Merak sub-pipelined TMP and Megatron-LM TMP scale well in our cluster. This is dut to the small gap in communication performance between inter- and intra-node, GPUs within node interconnect via PCIe links while nodes are connected via 100Gbps InfiniBand. Compared to high-speed links such as NVLink, the more commonly low intra-node bandwidth brings larger communication overheads on TMP, making the overlapped communications more necessary.

\section{Conclusion}

3D parallelism has become the SOTA training strategy for giant foundation deep neural networks.
To address the generality and inefficiency of 3D parallelism, we propose a user-friendly and high-performance distributed training framework called Merak. 
We design a non-intrusive API to ensure easy access to 3D parallelism for community models.
Merak also improves training performance by integrating three training optimization techniques in its efficient 3D parallel runtime engine.
Finally, the experimental results demonstrate that compared with the best baselines, Merak achieves up to 1.61$\times$ acceleration in training, maintains performance improvements on different scales, and produces considerable benefit for each optimization.
We have open-sourced implementation of Merak, and expect that the community will add support to existing and future works.

\ifCLASSOPTIONcompsoc
  \section*{Acknowledgments}
\else
  \section*{Acknowledgment}
\fi

The work was partially supported by the National Key R\&D Program of China (No. 2021YFB0301200) and National Natural Science Foundation of China (No. 62025208). Dongsheng Li is the corresponding author of this paper.

\ifCLASSOPTIONcaptionsoff
  \newpage
\fi


\bibliographystyle{IEEEtran}
\bibliography{sample}
%



%

\begin{IEEEbiography}[{\includegraphics[width=1in,height=1.25in,clip,keepaspectratio]{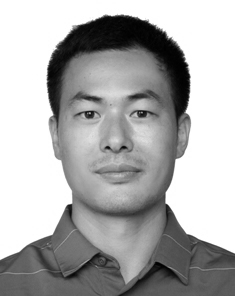}}]{Zhiquan Lai}
 received his Ph.D, M.S. and B.S. degrees in Computer Science from National University of Defense Technology (NUDT) in 2015, 2010 and 2008 respectively.
He is currently an associate professor in the National Key Laboratory for Parallel and Distributed Processing of NUDT.
He worked as a research assistant at Department of Computer Science, the University of Hong Kong during Oct. 2012 to Oct. 2013. His current research interests include high-performance system software, distributed machine learning, and power-aware computing.
\end{IEEEbiography}

\begin{IEEEbiography}[{\includegraphics[width=1in,height=1.25in,clip,keepaspectratio]{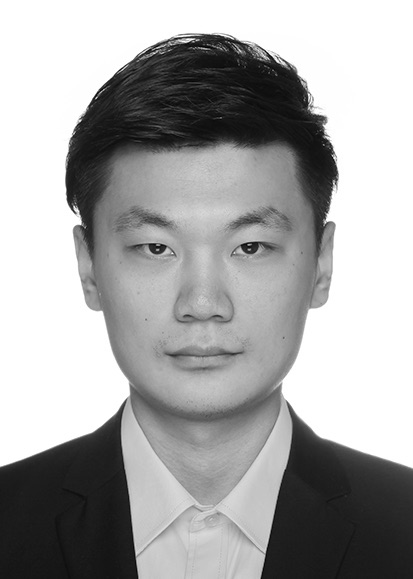}}]{Shengwei Li}
 received the B.S. degree from Nanjing University, Jiangsu, China in 2017, and the M.S. degree in computer science from Stony Brook University, New York, USA in 2020. He is pursuing his Ph.D. degree at the College of Computer, NUDT. His research interests include high-performance computing and distributed machine learning systems.
\end{IEEEbiography}

\begin{IEEEbiography}[{\includegraphics[width=1in,height=1.25in,clip,keepaspectratio]{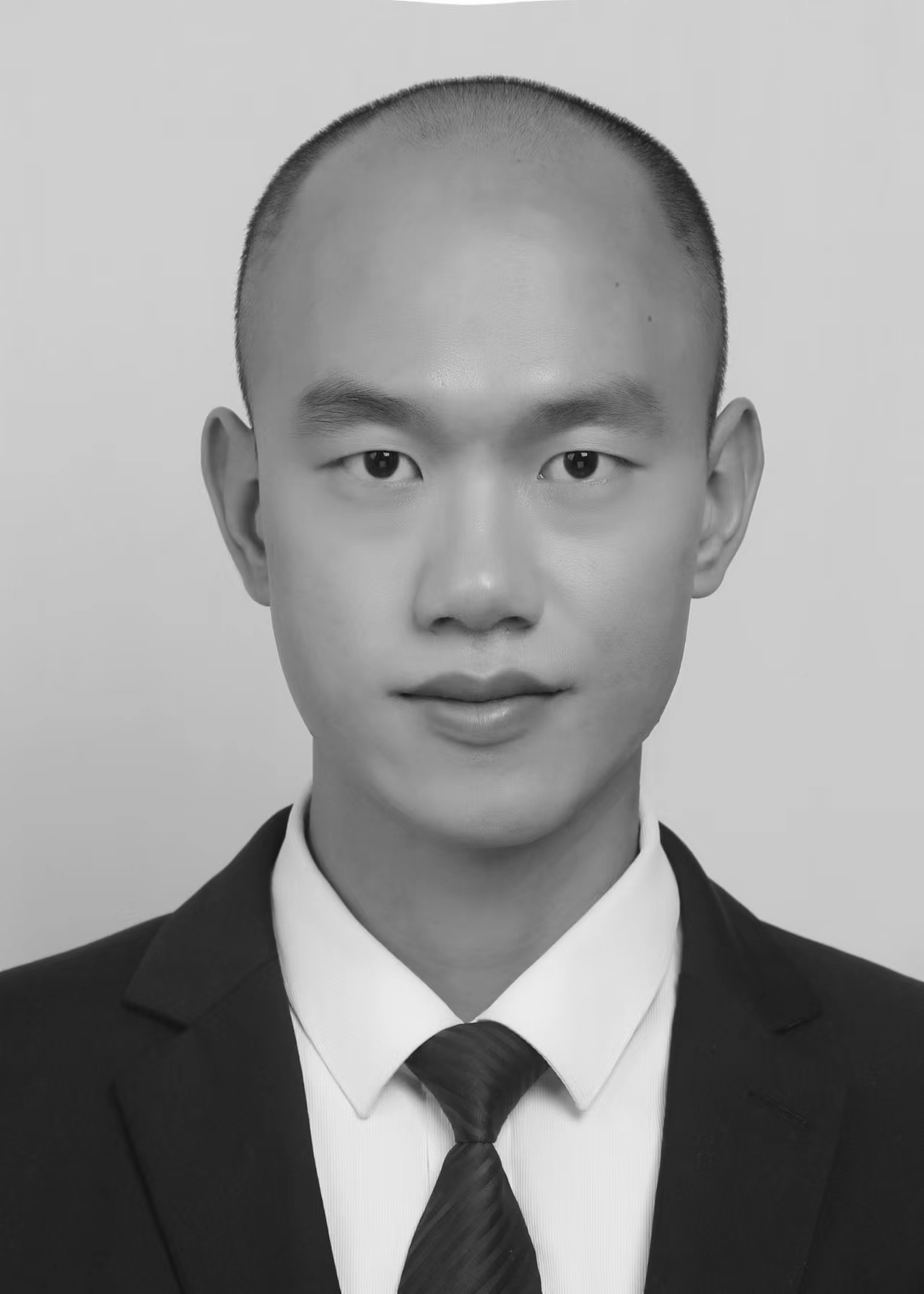}}]{Xudong Tang}
 received the Bachelor and Master degree in Xiangtan University, Hunan, China in July 2016 and July 2019, respectively. He is currently an engineer in the National Key Laboratory for Parallel and Distributed Processing of NUDT. His research interests include high-performance computing and distributed machine learning systems.
\end{IEEEbiography}

\begin{IEEEbiography}[{\includegraphics[width=1in,height=1.25in,clip,keepaspectratio]{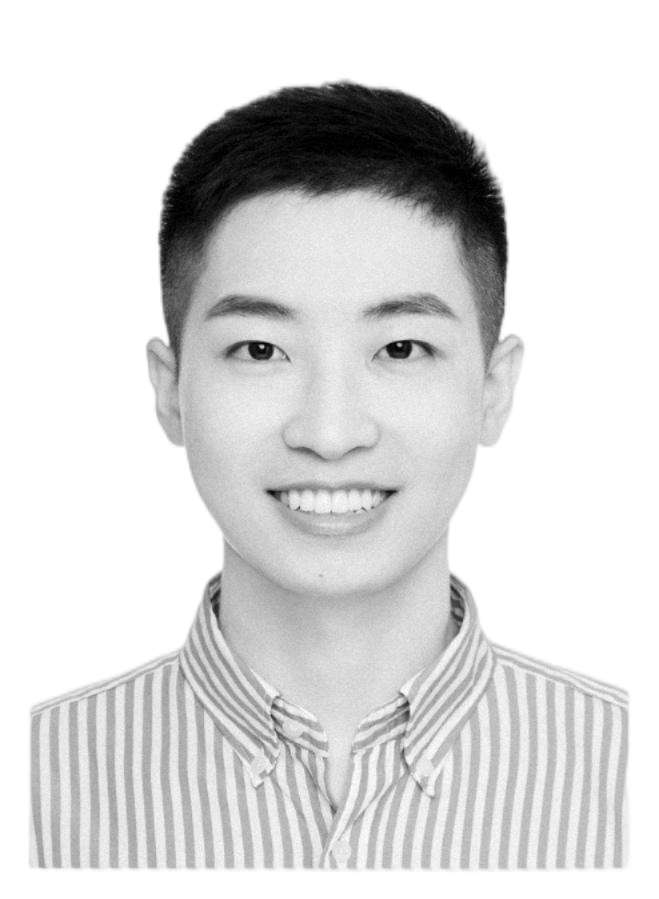}}]{Keshi Ge} received his B.S. degree from the Department of Computer Science and Technology, Xi’an Jiaotong University, China, in 2015, and his Ph.D. and M.S. degree from the College of Computer, National University of Defense Technology (NUDT), in 2022 and 2017, respectively. He worked as a visiting Ph.D. student at the Department of Electrical and Computer Engineering, University of Alberta, from Nov. 2019 to Aug. 2020. His research interests include high-performance computing and distributed machine learning systems.
\end{IEEEbiography}

\enlargethispage{-2in}
\vfill

\begin{IEEEbiography}[{\includegraphics[width=1in,height=1.25in,clip,keepaspectratio]{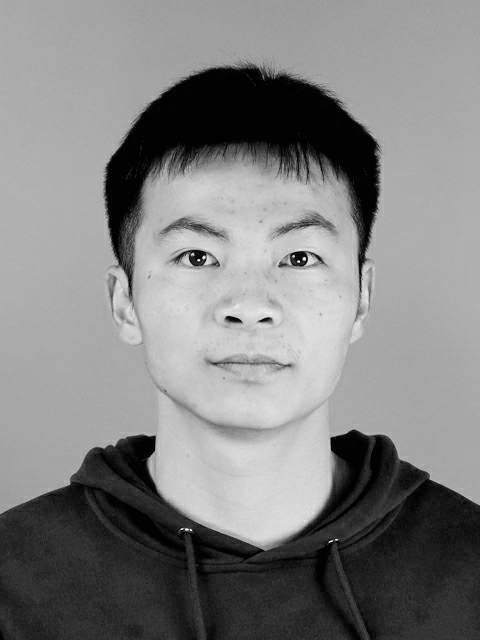}}]{Weijie Liu}
 received his Bachelor degree in computer science from Nankai University, China, in 2020, and his M.S. degree from the College of Computer, National University of Defense Technology (NUDT), in 2022. He is pursuing his Ph.D. degree at the College of Computer, NUDT. His current interests are mainly in optimization techniques related to large-scale model training.
\end{IEEEbiography}

\begin{IEEEbiography}[{\includegraphics[width=1in,height=1.25in,clip,keepaspectratio]{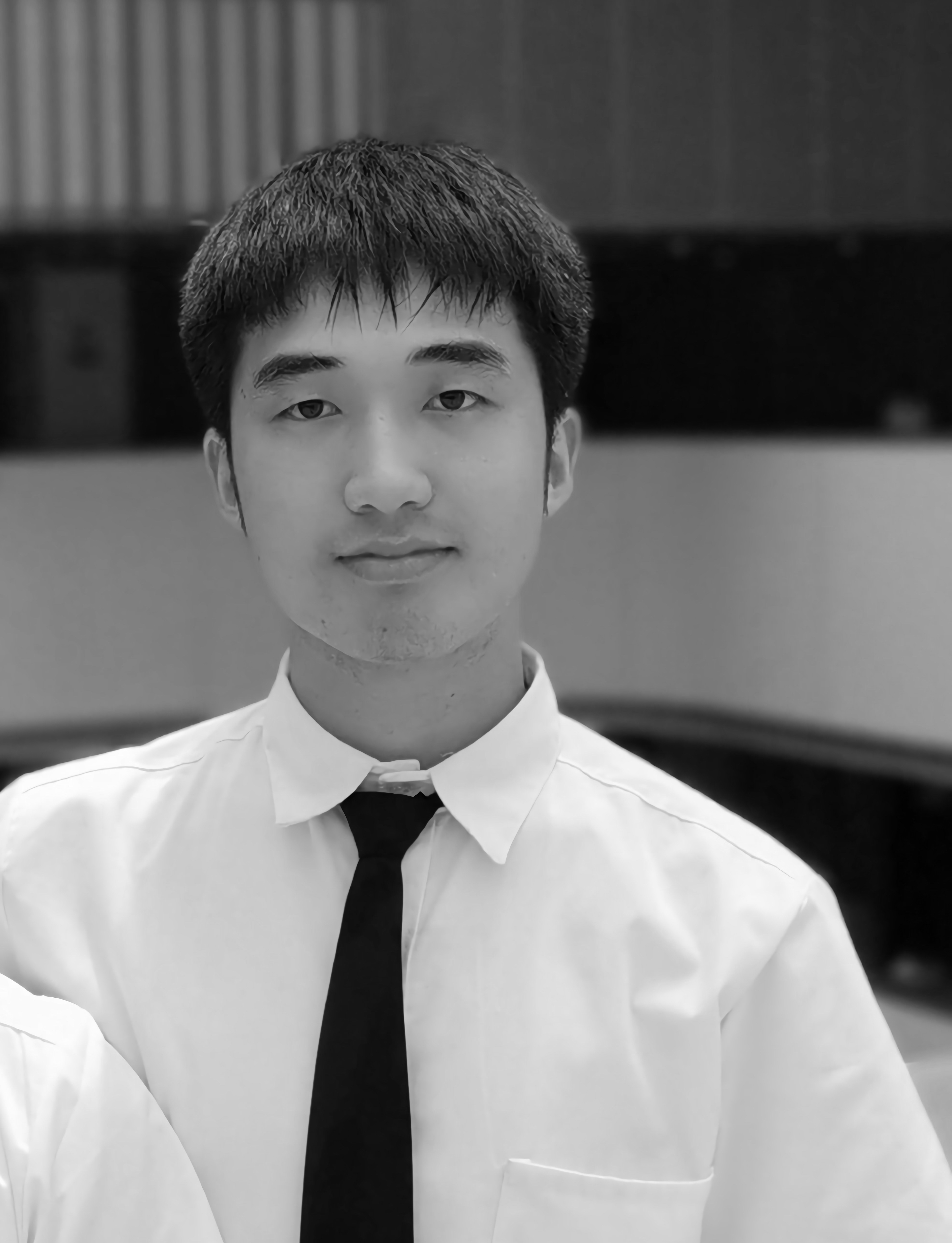}}]{Yabo Duan} 
received his Bachelor degree in computer science from Northwestern Polytechnical University, China, in 2020, and his M.S. degree from the College of Computer, National University of Defense Technology (NUDT), in 2022. His current interests are mainly in optimization techniques related to large-scale model training.
\end{IEEEbiography}

\begin{IEEEbiography}[{\includegraphics[width=1in,height=1.25in,clip,keepaspectratio]{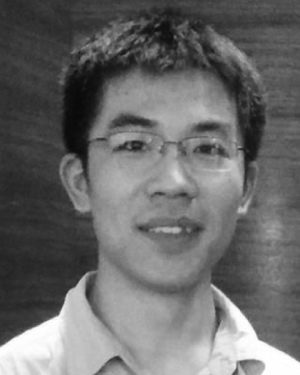}}]{Linbo Qiao} received the B.S., M.S., and Ph.D. degrees in computer science and technology from the National University of Defense Technology (NUDT), Changsha, China, in 2010, 2012, and 2017, respectively. He is currently an assistant professor with the Parallel and Distributed Processing Laboratory, NUDT. He worked as a research assistant in Chinese University of Hong Kong, from May 2014 to Oct. 2014. His research interests include structured sparse learning, online and distributed optimization, and deep learning for graph and graphical models.
\end{IEEEbiography}

\begin{IEEEbiography}[{\includegraphics[width=1in,height=1.25in,clip,keepaspectratio]{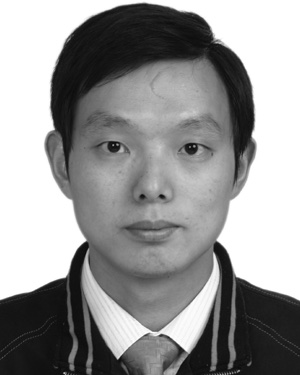}}]{Dongsheng Li}
 is a professor and doctoral supervisor in the College of Computer at National University of Defense Technology (NUDT). He received his PhD degree in computer science and technology from NUDT in 2005. He was awarded the Chinese National Excellent Doctoral Dissertation in 2008. His research interests include distributed systems, cloud computing and big data processing.
\end{IEEEbiography}


\vfill


\end{document}